\documentclass{article}

\PassOptionsToPackage{numbers, sort}{natbib}
\usepackage[preprint]{neurips_2026}
\makeatletter
\DeclareRobustCommand\onedot{\futurelet\@let@token\@onedot}
\def\@onedot{\ifx\@let@token.\else.\null\fi\xspace}

\def\eg{\emph{e.g}\onedot,~} 
\def\ie{\emph{i.e}\onedot,~}

\makeatother

\usepackage[utf8]{inputenc}
\usepackage[T1]{fontenc}
\usepackage{url}
\usepackage{booktabs}
\usepackage{amsmath}
\usepackage{amssymb}
\usepackage{amsfonts}
\usepackage{nicefrac}
\usepackage{microtype}
\usepackage{subcaption}
\usepackage[table]{xcolor}
\definecolor{refblue}{rgb}{0.21,0.49,0.74}
\usepackage[pagebackref,breaklinks,colorlinks,allcolors=refblue]{hyperref}

\usepackage{graphicx}
\graphicspath{{images/}}
\usepackage{xspace}
\usepackage{enumitem}
\usepackage{caption}
\usepackage{multirow}

\usepackage{algorithm}
\usepackage{algpseudocode}

\usepackage{nicematrix}

\usepackage[capitalize]{cleveref}
\crefname{section}{Section}{Sections}
\crefname{table}{Table}{Tables}
\crefname{figure}{Figure}{Figures}
\crefname{equation}{Equation}{Equations}
\crefname{appendix}{Appendix}{Appendices}

\usepackage{xfp}

\newcommand{\method}{{Dexterous Point Policy}\xspace}

\newcommand{\ppbottle}{0.0}
\newcommand{\ppbox}{0.0}
\newcommand{\ppball}{4.2}
\newcommand{\pptowel}{12.5}
\newcommand{\ppteddy}{4.2}
\newcommand{\ppopen}{8.4}
\newcommand{\ppbrush}{0.0}
\newcommand{\ppspray}{0.0}

\newcommand{\pppnpavg}{\fpeval{round((\ppbottle + \ppbox + \ppball + \pptowel + \ppteddy) / 5, 1)}}
\newcommand{\pptooluseavg}{\fpeval{round((\ppopen + \ppbrush + \ppspray) / 3, 1)}}
\newcommand{\ppavg}{\fpeval{round((\ppbottle + \ppbox + \ppball + \pptowel + \ppteddy + \ppopen + \ppbrush + \ppspray) / 8, 1)}}

\newcommand{\vitrabottle}{0.0}
\newcommand{\vitrabox}{0.0}
\newcommand{\vitraball}{0.0}
\newcommand{\vitratowel}{4.2}
\newcommand{\vitrateddy}{0.0}
\newcommand{\vitraopen}{4.2}
\newcommand{\vitrabrush}{0.0}
\newcommand{\vitraspray}{0.0}

\newcommand{\vitrapnpavg}{\fpeval{round((\vitrabottle + \vitrabox + \vitraball + \vitratowel + \vitrateddy) / 5, 1)}}
\newcommand{\vitratooluseavg}{\fpeval{round((\vitraopen + \vitrabrush + \vitraspray) / 3, 1)}}
\newcommand{\vitraavg}{\fpeval{round((\vitrabottle + \vitrabox + \vitraball + \vitratowel + \vitrateddy + \vitraopen + \vitrabrush + \vitraspray) / 8, 1)}.0}

\newcommand{\oursbottle}{95.8}
\newcommand{\oursbox}{75.0}
\newcommand{\oursball}{70.8}
\newcommand{\ourstowel}{87.5}
\newcommand{\oursteddy}{79.2}
\newcommand{\oursopen}{87.5}
\newcommand{\oursbrush}{62.5}
\newcommand{\oursspray}{41.7}

\newcommand{\ourspnpavg}{\fpeval{round((\oursbottle + \oursbox + \oursball + \ourstowel + \oursteddy) / 5, 1)}}
\newcommand{\ourstooluseavg}{\fpeval{round((\oursopen + \oursbrush + \oursspray) / 3, 1)}}
\newcommand{\oursavg}{\fpeval{round((\oursbottle + \oursbox + \oursball + \ourstowel + \oursteddy + \oursopen + \oursbrush + \oursspray) / 8, 1)}.0}

\newcommand{\vspp}{\fpeval{round((\oursavg - \ppavg), 1)}}

\title{Dexterous Point Policy: Learning Point-based Dexterous Hand Policies from Human Demonstrations}

\author{%
  \parbox{0.9\textwidth}{\centering\bfseries
  Beomjun Kim\thanks{Equal contribution} \quad
  Seong Hyeon Park\footnotemark[1] \quad
  Seunghoon Sim \quad
  Seungjun Moon \quad
  Sanghyeok Lee \quad
  Jinwoo Shin}\\[0.85em]
  KAIST \\[0.15em]
  \texttt{\{bullbum1126, seonghyp, jinwoos\}@kaist.ac.kr}
}

\begin{document}
\maketitle

\begin{abstract}
Robotic foundation models pre-trained on human demonstration videos have shown promise,
but a significant embodiment gap remains when the resulting policies are deployed on real robots.
A common remedy is to fine-tune these models on robot-specific demonstrations.
However, robot data collection can be prohibitively expensive and time-consuming, which is particularly acute in dexterous manipulation, \eg teleoperating a multi-fingered hand for even a single atomic task can take days.
To address this, we introduce \method{}, a framework that learns dexterous manipulation policies directly from human videos and requires no robot demonstrations.
Our core insight is that a unified 3D keypoint representation can bridge human and robot embodiments when used for both observations and actions.
Specifically, we extract 3D keypoints of task-relevant objects and human hands from raw videos, and train an autoregressive transformer over these keypoints.
We observe that at the keypoint level, specifically the wrist and fingertips, human and robot behaviors closely align, enabling direct policy transfer.
On a suite of real-robot tasks spanning pick-and-place and tool use, \method{} attains \oursavg\% success, whereas a state-of-the-art VLA baseline reaches only \vitraavg\%.
Furthermore, our method generalizes strongly to unseen scenarios, including multi-object environments and novel object categories.
\end{abstract}

\section{Introduction}
\label{sec:intro}

Data scaling has been one of the most important drivers of recent progress in artificial intelligence, powering the rise of general-purpose systems in language~\citep{achiam2023gpt,touvron2023llama,team2023gemini}, vision~\citep{ramesh2022dalle,rombach2022stablediffusion}, and video~\citep{brooks2024sora,yang2024cogvideox}. Robotics, by contrast, has yet to see its scaling moment. Robot trajectories cannot be scraped from the web; every datapoint must be physically executed through teleoperation~\citep{zhao2023aloha,khazatsky2024droid}. Even the largest open robot datasets~\citep{o2023openx,khazatsky2024droid} remain orders of magnitude smaller than their counterparts in language and vision, and assembling them takes months or years of dedicated human effort. This scarcity is especially acute for \emph{dexterous} manipulation: multi-finger hands have far higher action dimensionality than parallel-jaw grippers, making each demonstration slower, costlier, and harder to collect cleanly.

Recently, robotic foundation models choose to mine the vast pool of human videos available online~\citep{grauman2022ego4d,grauman2024egoexo4d,goyal2017something,damen2018scaling} as a scalable substitute for robot data. They learn visual or reward representations from human videos~\citep{nair2023r3m,ma2023vip,karamcheti2023language,majumdar2023vc1}, extract coarse action priors from hand motion~\citep{bahl2023affordance,bharadhwaj2024track2act,bharadhwaj2024gen2act,wen2023motiontracks}, or pretrain Vision-Language-Action (VLA) models on hand-action labels distilled from egocentric videos~\citep{bjorck2025gr00t,intelligence2025pi05,vitra2025}. However, these approaches encounter a common bottleneck: the significant embodiment gap between humans and robots is too wide at the pixel or raw-joint level to close from human data alone, so a substantial amount of in-domain robot teleoperation is still required at fine-tuning. The data-scaling that human videos promise on paper is therefore throttled in practice by the same robot-data bottleneck these methods were meant to alleviate.

\begin{figure}[t]
\centering
\includegraphics[width=\linewidth]{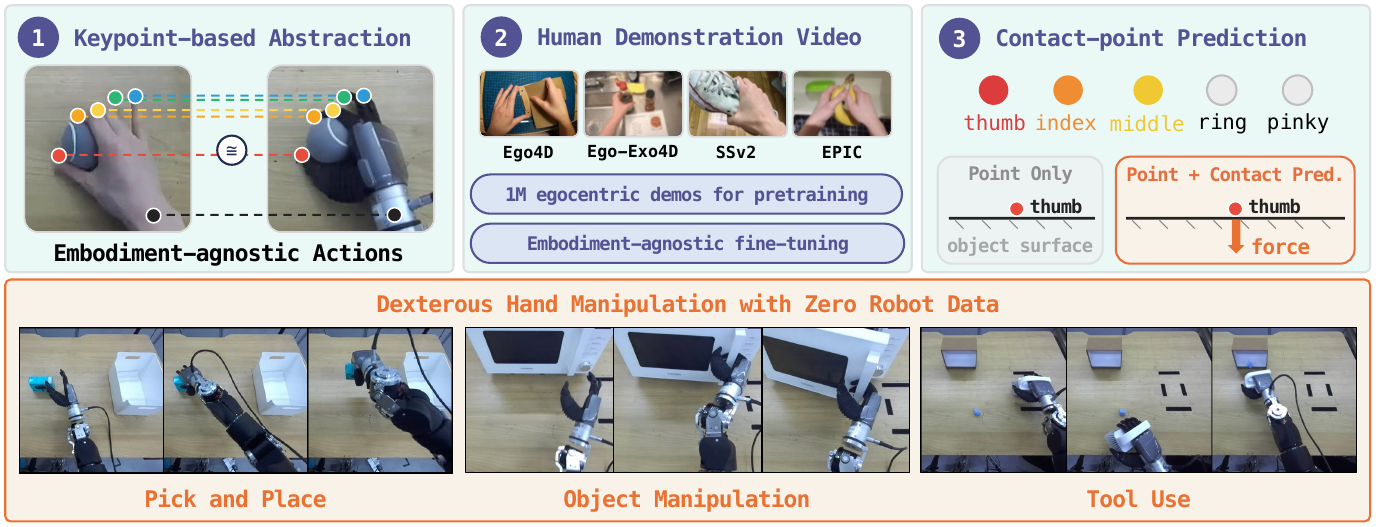}
\caption{\textbf{\method.} We present a dexterous manipulation policy trained solely from human demonstration videos.
Our method combines (1) a six-keypoint hand abstraction shared by human and robot, (2) internet-scale human-video pretraining and per-task fine-tuning, and (3) a fingertip contact prediction that injects force on top of the otherwise point-only representation. Together they enable a multi-finger robot to perform eight dexterous tasks, trained without any robot demonstrations.}
\label{fig:teaser}
\end{figure}

Meanwhile, there is another line of work that proposes keypoint-based representations as a way to sidestep the embodiment gap: by abstracting both the scene and the end-effector into a small set of task-relevant 3D points, policies become agnostic to visual appearance and to the fine-grained morphology of the agent~\citep{pointpolicy,pointbridge}. For example, Point Policy~\citep{pointpolicy} demonstrated that a gripper policy can be learned from human hand videos alone by aligning the two-finger gripper with the thumb and index fingertips of the human hand. While their results are compelling, they are fundamentally constrained by a gripper-centric representation. This limitation restricts learning to specialized human demonstrations where users must intentionally adopt unnatural, gripper-mimicking hand poses, thereby precluding these models from leveraging the vast diversity of general human videos available at internet scale.

In this work, we introduce \textbf{\method}, a framework that learns dexterous manipulation policies from human videos alone---with zero robot demonstrations at any stage of training---through a unified keypoint representation that both scales with internet-scale human video and transfers directly from human to robot. \method{} builds on three advances over prior point-based policies. (i) A \emph{six-keypoint hand abstraction} (wrist and five fingertips) for multi-finger coordination which provides a simple-yet-effective 3D keypoint representation shared by human and robot hands. (ii) A \emph{lightweight contact-point prediction}, jointly produced with the hand trajectory, that recovers the contact-force modality which point-only representations cannot express. (iii) \emph{Internet-scale human-video pretraining}: we pretrain an autoregressive transformer on the {VITRA} corpus~\citep{vitra2025} ($\sim 1$M egocentric episodes aggregated from \texttt{Ego4D}~\citep{grauman2022ego4d}, \texttt{Ego-Exo4D}~\citep{grauman2024egoexo4d}, \texttt{Something-Something~v2}~\citep{goyal2017something}, and \texttt{EPIC-KITCHENS}~\citep{damen2018scaling}) and fine-tune on a small number of task-specific human demonstrations.
At deployment, predicted hand keypoints are translated to joint targets via inverse kinematics, and predicted contact flags drive a small joint offset that injects the required fingertip force. No robot teleoperation, co-training, or fine-tuning is used.

We evaluate \method{} on a suite of dexterous manipulation tasks spanning pick-and-place and tool use, deployed on an {OpenArm} bimanual arm equipped with Inspire {RH56F1} dexterous hands. Our main findings are:
\begin{itemize}[leftmargin=1.5em,itemsep=2pt]
  \item \method{} is, to our knowledge, the first method to train a \emph{dexterous} manipulation policy with \emph{zero} robot demonstrations, achieving \oursavg\% success on real-robot tasks.
  \item The contact-point prediction mechanism, driven by minimal annotation, recovers the missing force modality of point representations and accounts for a +\vspp-point absolute improvement over a point-only baseline.
  \item Internet-scale human-video pretraining transfers cleanly through the unified keypoint representation, improving downstream success by +14.2 points over a fine-tune-only ablation.
\end{itemize}

By learning dexterous manipulation from human data alone, \method{} takes a concrete step toward addressing two major bottlenecks of robot policy learning: data scaling and human-to-robot transfer.

\section{Related Work}
\label{sec:related}

\subsection{Point-based Robot Policy}

Point-based representations have a long history in robotic manipulation, but have traditionally been placed on only one side of the policy: either as a compact observation consumed by an optimizer or neural policy that still acts in joint or end-effector space~\citep{manuelli2019kpam,simeonov2022ndf,huang2024rekep,liu2024moka,yuan2024robopoint,levy2024p3po}, or as an action prior in the form of point trajectories predicted from image observations~\citep{bharadhwaj2024track2act,bharadhwaj2024gen2act,wen2023motiontracks}. Point Policy~\citep{pointpolicy} and its extension Point Bridge~\citep{pointbridge} are, to our knowledge, the first to place points on \emph{both} sides---the policy observes and predicts 3D keypoints in a single Cartesian space shared by the human demonstrator and the robot end-effector---which is the formulation \method{} builds on.

\paragraph{Point Policy.}
Point Policy~\citep{pointpolicy} first proposed a keypoint-based unified observation and action space for robot policy learning, showing that gripper policies can be learned from human hand videos by relating the two-finger gripper to the thumb and index fingertips of the human hand---that is, using the point representation to mitigate the embodiment gap.

\paragraph{Point Bridge.}
Point Bridge~\citep{pointbridge} builds on the observation that keypoint-based robot policies are robust to the visual domain gap, and studies sim-to-real learning and zero-shot transfer in this setting---that is, using the point representation to mitigate the visual domain gap. It further proposes an automated pipeline that leverages recent VLMs and 3D priors (segmentation and metric depth estimation) to extract keypoints for task-relevant objects and the end-effector from a single-view image.

These works are limited to gripper control and do not leverage internet-scale video, leaving the data-scaling bottleneck open. We extend point policies to the dexterous-hand embodiment, making them applicable to more complex manipulation tasks, and propose a method for leveraging human hand data at scale, including internet-scale human videos.

\subsection{Robot Policy Learning from Human Videos}

Motivated by the difficulty of scaling robot data, learning robot policies from human videos has been actively studied in recent years. Several works leverage egocentric human videos to learn visual or visuomotor representations~\citep{nair2023r3m,ma2023vip,karamcheti2023language,majumdar2023vc1}. Others use explicit human actions extracted from mocap videos or web videos to guide imitation learning~\citep{mandikal2022dexvip,shaw2023videodex}. A third line learns affordances, point trajectories, or hand-object masks from human videos to guide policy learning~\citep{bahl2023affordance,bharadhwaj2024track2act,bharadhwaj2024gen2act,wen2023motiontracks}. Another group learns latent actions from human videos in an unsupervised manner and pretrains action models on latent action labels, while more recent methods extract 3D hand actions from egocentric videos for VLA pretraining~\citep{bjorck2025gr00t,intelligence2025pi05}. A separate direction uses human videos to train video generation models, visual task planners, or world models for manipulation.

\paragraph{Phantom.}
A separate effort closes the human-to-robot gap at the \emph{image} level rather than the representation level: Phantom~\citep{lepert2025phantom} collects videos of a human arm performing the task and inpaints the gripper appearance in place of the human hand, enabling training without robot demonstrations under a 2D image input and end-effector pose action policy. This is complementary to our representation-level approach but is restricted to gripper end-effectors.

\paragraph{VITRA.}
VITRA~\citep{vitra2025} trains a Vision-Language-Action model on internet-scale human hand video. It defines actions in joint space and treats the robot hand action space as a subspace of the human hand action space. Fine-tuning nonetheless requires robot demonstrations---$2610$ in their setting---so data scaling and human-to-robot transfer are only partially addressed.

In contrast, \method{} trains a \emph{dexterous} manipulation policy from human videos \emph{alone}, using a unified keypoint representation to reduce the embodiment gap and remove the need for robot demonstrations during training.

\section{\method{}}
\label{sec:method}

\method{} uses a unified observation and action representation that reduces the alignment required for human-to-robot policy transfer, incorporates internet-scale human video pretraining, and also fine-tunes from human videos only. An overview is provided below, with details in the following sections.

\subsection{Overview}
\label{subsec:overview}
\method{} learns a dexterous manipulation policy from human videos alone through a unified 3D-keypoint representation: both task-relevant objects and a six-keypoint hand abstraction (wrist and five fingertips) are extracted from each frame using off-the-shelf VLM, segmentation, depth, and hand-tracking models (\cref{subsec:data}), and the same representation is shared by the human demonstrator and the robot end-effector. An autoregressive transformer is pretrained on $\sim 1$M egocentric human episodes and then fine-tuned on a small set of task-specific human demonstrations augmented with sparse per-frame fingertip contact annotations; at the input, the contact annotation is fused into the hand token, and at the output, contact is predicted jointly with the next-step hand trajectory (\cref{subsec:policy_learning}). At deployment, predicted keypoints are mapped to robot joint targets via inverse kinematics, and predicted contact flags inject a small fingertip joint offset that supplies the contact force a point-only representation cannot express (\cref{subsec:inference}).

\subsection{Data}
\label{subsec:data}
\begin{figure}[t]
    \centering
    \includegraphics[width=0.95\linewidth]{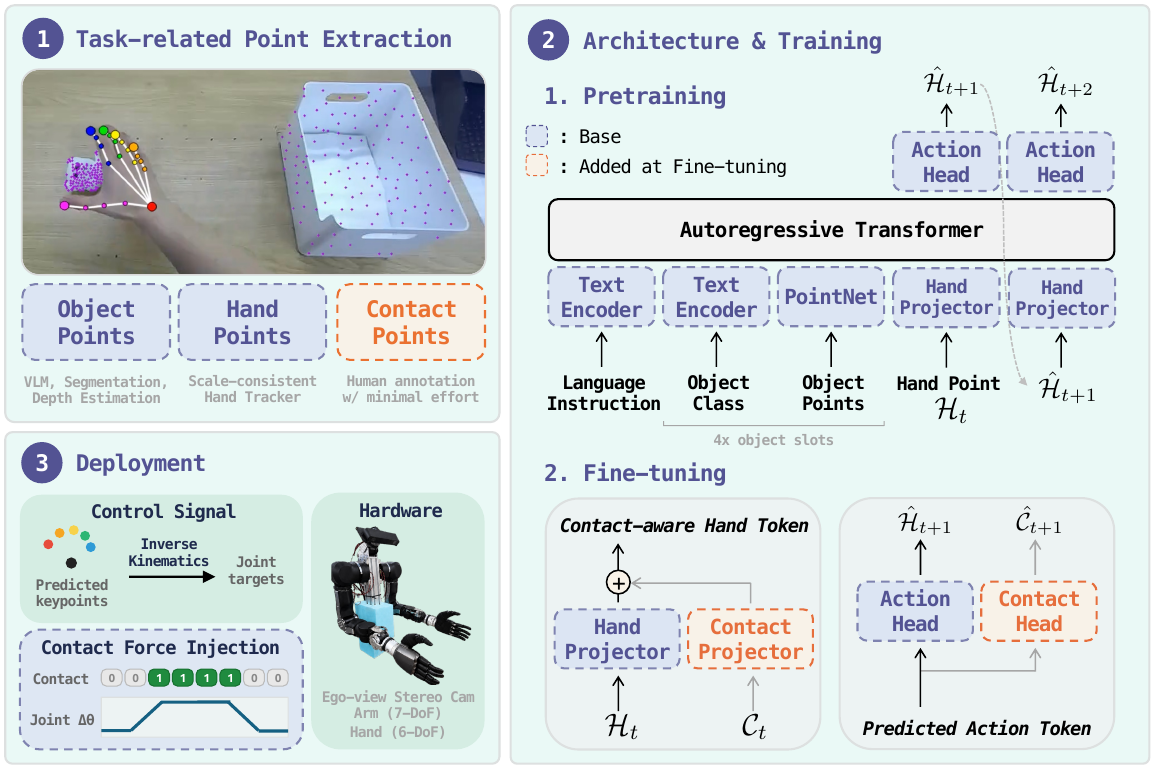}
    \caption{\textbf{Overview.} (1) Point Extraction: From an egocentric frame and task description, we extract \textit{object points} (via VLM segmentation and depth estimation), \textit{hand points} (wrist and five fingertips from a hand tracker), and \textit{contact points} (lightweight manual annotation). (2) Architecture \& Training: During \textit{pretraining}, an autoregressive transformer takes language, object names, object points, and the current hand point $\mathcal{H}_t$ to predict future hand points $\hat{\mathcal{H}}_{t+1}$. During \textit{fine-tuning}, a contact projector fuses contact flags $\mathcal{C}_t$ with the hand token, and an added contact head jointly predicts $\hat{\mathcal{H}}_{t+1}$ and $\hat{\mathcal{C}}_{t+1}$. (3) Deployment: Predicted keypoints map to joint targets via inverse kinematics, and predicted contacts modulate trajectories through contact-force injection on the robot hardware.}
    \label{fig:point_extraction}
\end{figure}

\paragraph{Human video collection.}
\method{} uses two sources of human video. Our \emph{pretraining corpus} is the {VITRA} dataset~\citep{vitra2025}, which aggregates four egocentric human-video collections: \texttt{Ego4D}~\citep{grauman2022ego4d}, \texttt{Ego-Exo4D}~\citep{grauman2024egoexo4d}, \texttt{Something-Something~v2}~\citep{goyal2017something}, and \texttt{EPIC-KITCHENS}~\citep{damen2018scaling}, and per-frame language captions and hand keypoints extracted from the pretrained \texttt{HaWoR} hand tracker~\citep{hawor2024}\footnote{We utilize the pre-processed captions available in the open source repository of VITRA \citep{vitra2025}.}.
In total this gives approximately $1\mathrm{M}$ episodes and $240$ hours of video. Our \emph{fine-tuning set} is collected per downstream task: because only an ego-view camera and the operator's bare hand are required, a single worker collects roughly $200$ demonstrations per hour, substantially faster than teleoperation. Per task we collect $500$ demonstrations ($\sim 1.2$ hours of video, $\sim 3$ hours of single-worker effort).

We apply our point-extraction pipeline, depicted in \cref{fig:point_extraction}, to both the pretraining and fine-tuning corpus, which produces two streams: (A) 3D points of task-relevant objects and (B) the six-keypoint hand abstraction. For the pre-training corpus the {VITRA} dataset already supplies (B) and the language captions; we obtain (A) by running the object-point pipeline below on its frames. For the fine-tuning set we run both (A) and (B) ourselves on the freshly collected videos, and additionally collect contact annotations.

\paragraph{Object-point extraction.}
Given an initial scene image $\mathcal{I}_0$ and a natural language task description $\mathcal{L}$, we use \texttt{Qwen3.5-VL-8B-Instruct}~\citep{bai2025qwen3vl} to identify the set of task-relevant objects $\{l^1, \ldots, l^k\}$. For example, given the command ``pick the bottle and place it on the bowl,'' the model returns $\{\text{bottle}, \text{bowl}\}$. We then employ \texttt{SAM3}'s text-query-based object segmentation~\citep{meta2025sam3} to extract 2D segmentation masks $\{m_0^1, \ldots, m_0^k\}$, propagating them through subsequent frames using \texttt{SAM3}'s built-in memory for robust tracking under occlusion. For each timestep $t$, we sample $128$ 2D points uniformly from each mask $m_t^i$ and lift them to 3D using a depth map of the scene: \texttt{Depth-Anything-3}~\citep{depthanything3} for the pretraining corpus, where only single-view egocentric video is available; and stereo depth from the ego-view {ZED} camera on our robot embodiment for the fine-tuning set, since stereo is generally less noisy than monocular metric depth. Camera intrinsics are obtained per episode: from each source dataset's bundled metadata for the pretraining corpus, and from the ZED SDK calibration for the fine-tuning set. A transformation by camera extrinsics yields the final 3D point set $\mathcal{P}^{\mathrm{3D}}_t$ in world coordinates.

\paragraph{Hand-point extraction.}
We represent the dexterous end-effector, both human and robot, as a set of six keypoints: one at the wrist and one at each of the five fingertips. For the pretraining corpus we use VITRA's pre-computed \texttt{HaWoR}~\citep{hawor2024} hand keypoints directly.
Since the purpose of the pretraining is to coarsely train general dexterous hand manipulation, the less accurate keypoint labels are benign for training.
On the other hand, keypoints of the fine-tuning dataset are directly related to the task manipulation, which need to be accurate to successfully represent the process of the task.
Although \texttt{HaWoR} accurately tracks the hand pose itself, it infers hand shape and scale inconsistently even in a single sequence, which introduces a scale-depth ambiguity.
To address this, we introduce a modification to \texttt{HaWoR} to enable inference with a consistent hand scale.
This effectively resolves the scale-depth ambiguity and leads to a noticeable improvement in the stability of the keypoint trajectory.
We elucidate the details of scale-consistent \texttt{HaWoR} in \cref{app:hand-tracker}.

\paragraph{Contact-point annotation.}
A purely point-based policy predicts hand positions but carries no information about the contact force that must be applied when the hand meets an object. More concretely, once contact is established, hand and object keypoints stop moving even as the hand continues to apply force, so a point-only observation cannot distinguish a light touch from a firm grasp. For the multi-finger hand we use, object contact is mainly concentrated at the fingertips. Building on this, we add a lightweight annotation, \emph{at fine-tuning time only}: for each timestep, an annotator labels which fingertips are in contact with the target object, yielding a binary $5$-vector $c_t \in \{0,1\}^5$ indexed as $[\text{thumb}, \text{index}, \text{middle}, \text{ring}, \text{pinky}]$. The annotation adds roughly $10$ seconds per demonstration, so its overhead relative to the rest of the data collection is negligible. Pretraining uses no contact channel, as the VITRA corpus lacks contact labels. In practice, contact is a low-dimensional signal that can be readily inferred from fingertip-object proximity, and we observe that the policy learns it sufficiently during fine-tuning.

\subsection{Policy Training}
\label{subsec:policy_learning}

We instantiate the policy as an autoregressive transformer with two training stages: \emph{pretraining} on the {VITRA} corpus, in which only the hand and object channels are supervised, and \emph{fine-tuning} on the per-task human demonstrations, in which the contact channel is added on both the input and output sides.

\paragraph{Pretraining.}
At each timestep $t$ the input consists of four streams: a language instruction, the 3D points of task-relevant objects, the six hand keypoints, and the ego-view camera extrinsics. The language instruction is encoded by a {Sentence Transformer}~\citep{reimers2019sentencebert} into a single token. Each task-relevant object contributes two tokens: a \emph{semantic} token, obtained by encoding the VLM-returned object name with the same Sentence Transformer, and a \emph{geometry} token, obtained by encoding its 3D points with {PointNet}~\citep{qi2017pointnet}. The semantic token is placed immediately before its geometry token so that the transformer can bind points to object identities. We cap the number of objects per task at four (eight object tokens in total) and fill unused slots with a zero embedding the policy learns to ignore. The six hand keypoints share a fixed index (wrist, thumb, index, middle, ring, pinky); we concatenate their coordinates into an $18$-dim vector and project it to the model dimension via a hand projector $\phi_{\mathrm{hand}}$. Finally, we append the ego-view camera extrinsics as one additional token. Conditioned on this tokenized observation, the transformer autoregressively predicts a horizon of $H$ future hand positions, with each predicted step projected back as the input hand token for the next step (teacher forcing during training, own predictions at inference). At each step an action head $\psi_{\mathrm{act}}$ outputs six 3D hand keypoints $\hat{\mathcal{H}}_{t+h} \in \mathbb{R}^{6 \times 3}$ in the world frame, supervised with $l_1$ loss averaged over the batch, horizon, and keypoints:
\begin{equation}
\mathcal{L}_{\mathrm{act}} \;=\; \frac{1}{B\,H\,K}\sum_{b=1}^{B}\sum_{h=1}^{H}\sum_{k=1}^{K}\,\text{}\ell_1\!\left(\hat{\mathcal{H}}^{(b)}_{t+h,k} - \mathcal{H}^{(b)}_{t+h,k}\right),
\label{eq:loss_act}
\end{equation}
with $K=6$ keypoints and $H=16$ at pretraining. All four input streams are populated at pretraining time: language captions and hand keypoints are supplied by the VITRA dataset, object tokens are computed by running the \cref{subsec:data} pipeline on VITRA frames, and extrinsics are passed through. The contact channel is the sole channel inactive at this stage, since the corpus has no contact labels.

\paragraph{Fine-tuning.}
Fine-tuning uses $H=16$, initializes from the pretrained weights, and introduces the contact channel on both the input and output sides. At the input, a two-layer MLP contact projector $\phi_{\mathrm{contact}}$ maps the binary contact annotation $c_t \in \{0,1\}^5$ into the model dimension and \emph{adds} it to the hand embedding, forming a single contact-aware hand token; $\phi_{\mathrm{contact}}$'s last linear layer is zero-initialized so that the pretrained hand token is recovered exactly at the start of fine-tuning. At the output, a contact head $\psi_{\mathrm{ct}}$, added in parallel to the action head with default initialization, produces five logits whose sigmoid yields the per-fingertip contact probability $\hat{p}_{t+h} \in [0,1]^5$. The fine-tuning objective is
\begin{equation}
\mathcal{L}_{\mathrm{ft}} \;=\; \mathcal{L}_{\mathrm{act}} + \lambda\,\mathcal{L}_{\mathrm{ct}},
\qquad
\mathcal{L}_{\mathrm{ct}} \;=\; \mathrm{BCE}_{w^{+}}(\hat{p}, c),
\label{eq:loss_ft}
\end{equation}
with $\mathcal{L}_{\mathrm{act}}$ as in \eqref{eq:loss_act}, $\lambda = 1$, and $w^{+}$ a positive-class weight that compensates for class imbalance. We stop the contact-loss gradient at the transformer backbone: only $\psi_{\mathrm{ct}}$ receives gradients from $\mathcal{L}_{\mathrm{ct}}$, while $\phi_{\mathrm{contact}}$ is trained by $\mathcal{L}_{\mathrm{act}}$ flowing backward through the shared hand token. This decoupling keeps contact prediction from distorting the trajectory objective, yet still allows the input-side fusion to exploit contact labels to improve trajectory prediction---empirically lowering hand-trajectory loss relative to a trajectory-only baseline (\cref{subsec:exp_main}).

\subsection{Policy Inference}
\label{subsec:inference}

At deployment, the initial scene image $\mathcal{I}_0$ and task instruction $\mathcal{L}$ are used to obtain task-relevant objects $\{l^1, \ldots, l^k\}$ via the VLM, and their 2D keypoints $\mathcal{P}^{\mathrm{2D}}_0$ via \texttt{SAM3}; these are lifted to 3D using stereo depth from the ZED camera and the camera intrinsics. Robot hand points are computed from the robot URDF and its current joint state via forward kinematics. Because we use a static camera for all downstream tasks, we place the world frame at the camera frame and pass identity extrinsics to the policy. Mapping human hand motion onto a robot hand is itself an active research topic (motion retargeting). In our setting we sidestep this step: both human and robot hands are described by the same six-keypoint abstraction (wrist and five fingertips), so a trajectory predicted from human data is directly interpretable as a robot target. The predicted hand keypoints are translated to robot joint targets by a position-only damped least-squares IK solver that tracks the wrist and five fingertip sites on the robot URDF. The predicted contact logits are passed through a sigmoid and thresholded to a binary per-finger grip bit; when active, a per-joint closing offset is smoothly ramped in so that contact force is applied gradually rather than instantaneously. The resulting joint trajectory is executed by the robot controller at $20$\,Hz.

\section{Experiments}
\label{sec:exp}

\begin{figure}[t]
    \centering
    \includegraphics[width=\linewidth]{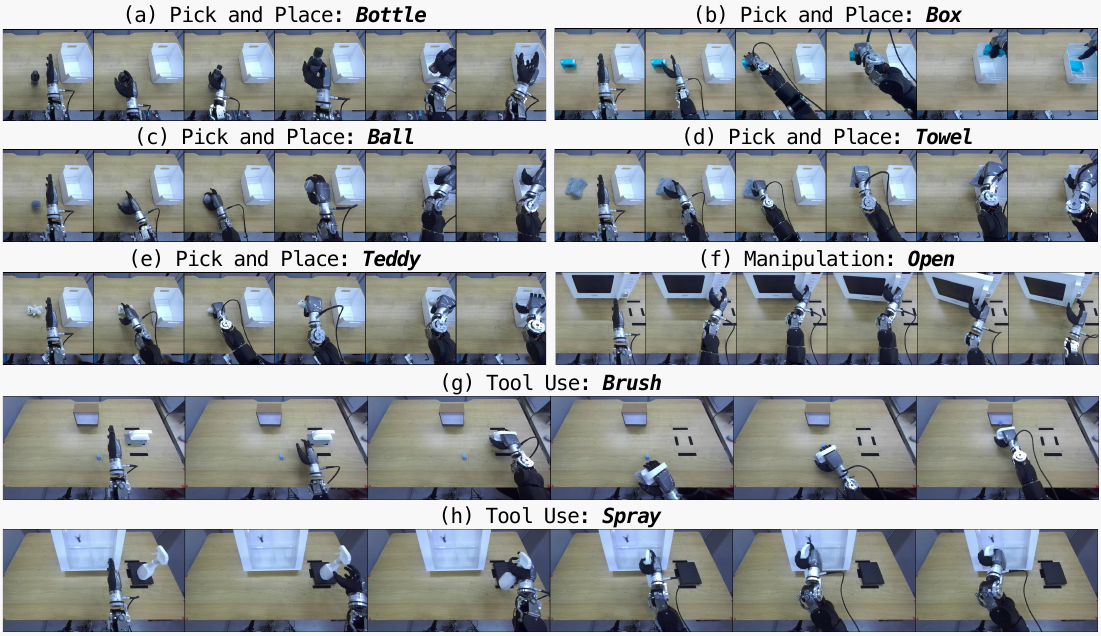}
    \caption{\textbf{Real-world rollouts.} Successful executions of \method{} across our suite of dexterous manipulation tasks, deployed on an OpenArm bimanual arm equipped with Inspire RH56F1 hands. All policies are trained from human videos alone, with no robot demonstrations.}
    \label{fig:rollout}
\end{figure}

Our experiments are designed to answer the following questions:
\begin{itemize}[leftmargin=1.5em,itemsep=2pt]
  \item \textbf{Q1.} Can \method{} succeed in various real-world dexterous manipulation tasks without any robot demonstrations?
  \item \textbf{Q2.} Can \method{} generalize to cluttered scenes and novel objects?
  \item \textbf{Q3.} Does the autoregressive modeling design improve the performance of the policy?
  \item \textbf{Q4.} Does internet-scale human-video pretraining improve downstream task success?
\end{itemize}

\subsection{Experimental Setup}
\label{subsec:exp_setup}

\paragraph{Hardware.}
We deploy on an {OpenArm} bimanual arm equipped with two Inspire {RH56F1} dexterous hands. A {ZED~2i} stereo camera, mounted to approximate the egocentric viewpoint of the human demonstrations, provides the robot's on-board perception at $1280\times720$ resolution and $30$\,fps; its built-in stereo depth map serves as the depth signal at both fine-tuning and deployment.

\paragraph{Tasks.}
We evaluate on two categories of real-world dexterous manipulation tasks:
\begin{itemize}[leftmargin=1.5em,itemsep=2pt]
  \item \textbf{Pick and Place.} The robot grasps an object from the table and places it in a fixed-position container. We use five objects spanning four shape categories: cylindrical (\emph{bottle}), spherical (\emph{tennis ball}), rectangular (\emph{box}), and deformable/irregular (\emph{towel}, \emph{teddy bear}). We collect $100$ human demonstrations per object, $500$ in total, and train a \emph{single} policy. The object's initial position is randomized across four predefined locations on the table.

  \item \textbf{Manipulation and Tool Use.} We evaluate three tasks that require hand--object interaction beyond pick-and-place: \emph{Open} (grasp the microwave handle and pull its door open),  \emph{Spray} (grasp the spray bottle and depress the trigger to a target position), and \emph{Brush} (grasp a hand brush and sweep debris to a target position). For each task, we collect $100$ human demonstrations and train a separate policy.
  
\end{itemize}
Further details on task setup and success criteria are provided in \cref{app:setup,app:success}.

\paragraph{Baselines.}
We compare \method{} against two baselines, both trained without robot demonstrations:
\begin{itemize}[leftmargin=1.5em,itemsep=2pt]
  \item \textbf{Point Policy}~\citep{pointpolicy}. We replace the original two-finger gripper action space with our six-keypoint hand representation (\cref{subsec:data}), keeping the original architecture and non-autoregressive action head unchanged.
  \item \textbf{VITRA}~\citep{vitra2025}. A Vision-Language-Action model pretrained on internet-scale human hand video. We adapt it to our setting by retargeting human-hand keypoint trajectories to robot joint trajectories via the same IK solver used at \method{}'s deployment (\cref{subsec:inference}), and use the resulting joint trajectories as fine-tuning supervision.
\end{itemize}
All methods are trained on the same human demonstration data. Full baseline details are in \cref{app:baselines}.

\paragraph{Evaluation protocol.}
For Pick and Place, we evaluate each object at all four initial positions with six trials per position, yielding $24$ trials per object and $120$ trials in total. For each Manipulation and Tool Use task, we also evaluate $24$ trials with randomized initial configurations. We report single-attempt success rate (\%).

\subsection{Dexterous Manipulation}
\label{subsec:exp_main}

\cref{tab:main} reports single-attempt success rates across all eight tasks.

\paragraph{Human-to-robot transfer (Q1).}
\method{} achieves an average success rate of $\ourspnpavg$\% on Pick and Place and $\ourstooluseavg$\% on Manipulation and Tool Use. Per-object results on Pick and Place are: bottle $\oursbottle$\%, box $\oursbox$\%, tennis ball $\oursball$\%, towel $\ourstowel$\%, and teddy bear $\oursteddy$\%.

\paragraph{Comparison with baselines.}
Both baselines, despite being trained on the same human demonstration data, struggle to transfer to the robot embodiment: {Point Policy} achieves $\pppnpavg$\% on Pick and Place and $\pptooluseavg$\% on Manipulation and Tool Use, while {VITRA} achieves $\vitrapnpavg$\% and $\vitratooluseavg$\%, respectively.
\method{} substantially outperforms both across all eight tasks, and \cref{fig:rollout} shows representative successful rollouts.
We observe that the remaining failures are dominated by low-level motor issues---imprecise action targeting and insufficient contact force---rather than higher-level task misunderstandings.

\begin{table}
\centering
\small
\caption{\textbf{Dexterous manipulation results.}
    We report the success rate (\%) over 24 trials per task on real-robot dexterous manipulation tasks. \textbf{Bold} indicates the best result.
}\label{tab:main}
\resizebox{\textwidth}{!}{
\begin{NiceTabular}{lccccccccc}
    \toprule
    & \multicolumn{5}{c}{Pick and Place} & \multicolumn{3}{c}{Manipulation \& Tool Use} \\
    \cmidrule(lr){2-6} \cmidrule(lr){7-9}
    Method & Bottle & Box & Ball & Towel & Teddy & Open & Brush & Spray & Avg. \\
    \midrule
    Point Policy
      & \ppbottle & \ppbox & \ppball & \pptowel & \ppteddy
      & \ppopen & \ppbrush & \ppspray
      & \ppavg \\
    VITRA
      & \vitrabottle & \vitrabox & \vitraball & \vitratowel & \vitrateddy
      & \vitraopen  & \vitrabrush & \vitraspray
      & \vitraavg \\
    \midrule
    \rowcolor{green!10}
    \textbf{DPP (Ours)}
      & \textbf{\oursbottle} & \textbf{\oursbox} & \textbf{\oursball} & \textbf{\ourstowel} & \textbf{\oursteddy} & \textbf{\oursopen}
     & \textbf{\oursbrush} & \textbf{\oursspray} & \textbf{\oursavg} \\
    \bottomrule
\end{NiceTabular}
}
\end{table}

\subsection{Generalization}
\label{subsec:exp_generalization}

We test generalization to cluttered scenes and unseen objects on the Pick and Place task (Q2).

\paragraph{Multi-object.}
In training, each demonstration involves a single object on an otherwise empty table. At test time, we place objects at all four initial positions simultaneously and instruct the robot to pick a designated target. \cref{tab:generalization} (top) reports success rates under this multi-object protocol, with $24$ trials per object. \method{} achieves $80.0$\% average success, compared to $\ourspnpavg$\% in the single-object setting. Results for all methods are reported in \cref{tab:generalization}.

\paragraph{Novel object.}
For each training object, we select one or two visually distinct novel objects that share the broad shape category but differ in size, color, and surface texture. We maintain $24$ trials per training category ($4$ positions $\times$ $6$ trials): when a category has one novel object, that object receives all six trials per position; when it has two, each receives three trials per position, yielding $120$ trials in total. \cref{tab:generalization} (bottom) reports the results. \method{} achieves 76.7\% on novel objects, compared to $\ourspnpavg$\% on training objects.

\begin{table}
\centering
\small
\caption{\textbf{Generalization on Pick and Place.}
    Success rate (\%) under multi-object and novel-object evaluation. \textbf{Bold} indicates the best result.
}\label{tab:generalization}
\begin{NiceTabular}{ll cccccc}
    \toprule
    & Method & Bottle & Box & Ball & Towel & Teddy & Avg. \\
    \midrule
    \multirow{3}{*}[-0.5\dimexpr\aboverulesep+\belowrulesep+\cmidrulewidth]{Multi-object~~}
    & Point Policy & 0.0 & 0.0 & 8.3 & 8.3 & 4.2 & 4.2 \\
    & VITRA        & 0.0 & 0.0 & 0.0 & 0.0 & 0.0 & 0.0 \\
    \addlinespace[-2pt] \cmidrule{2-8}
    & \cellcolor{green!10}\textbf{DPP (Ours)} & \cellcolor{green!10}\textbf{95.8} & \cellcolor{green!10}\textbf{70.8} & \cellcolor{green!10}\textbf{79.2} & \cellcolor{green!10}\textbf{83.3} & \cellcolor{green!10}\textbf{70.8} & \cellcolor{green!10}\textbf{80.0} \\
    \midrule
    \multirow{3}{*}[-0.5\dimexpr\aboverulesep+\belowrulesep+\cmidrulewidth]{Novel object~~}
    & Point Policy & 0.0 & 0.0 & 4.2 & 12.5 & 0.0 & 3.3 \\
    & VITRA        & 0.0 & 0.0 & 0.0 & 0.0 & 0.0 & 0.0 \\
    \addlinespace[-2pt] \cmidrule{2-8}
    & \cellcolor{green!10}\textbf{DPP (Ours)} & \cellcolor{green!10}\textbf{95.8} & \cellcolor{green!10}\textbf{75.0} & \cellcolor{green!10}\textbf{62.5} & \cellcolor{green!10}\textbf{87.5} & \cellcolor{green!10}\textbf{62.5} & \cellcolor{green!10}\textbf{76.7} \\
    \bottomrule
\end{NiceTabular}
\end{table}
\begin{table}
\centering
\small
\caption{\textbf{Ablation study.}
    Effect of removing individual components from DPP. We report success rate (\%) on Pick and Place tasks. \textbf{Bold} indicates the best result.
}\label{tab:ablation}
\begin{NiceTabular}{lcccccc}
    \toprule
    Method & Bottle & Box & Ball & Towel & Teddy & Avg. \\
    \midrule
    w/o AR       & 45.8 & 33.3 & 29.2 & 37.5 & 41.7 & 37.5 \\
    w/o Pretrain & 91.7 & 54.2 & 58.3 & 70.8 & 62.5 & 67.5 \\
    \midrule
    \rowcolor{green!10}
    \textbf{DPP (Full)} & \textbf{95.8} & \textbf{75.0} & \textbf{70.8} & \textbf{87.5} & \textbf{79.2} & \textbf{81.7} \\
    \bottomrule
\end{NiceTabular}
\end{table}

\subsection{Ablation Study}
\label{subsec:exp_ablation}

We ablate two design choices of \method{} and evaluate on the full task suite. Results are reported in \cref{tab:ablation}.

\paragraph{Autoregressive modeling (Q3).}
We replace \method{}'s autoregressive rollout across the $H$-step action chunk with a non-causal transformer that emits the full chunk in a single forward pass via a parallel-decoding MLP head, while keeping the contact head and the VITRA pretraining stage unchanged. This ablation (\emph{w/o AR}) achieves 37.5\% on Pick and Place, compared to 81.7\% for the full model.

\paragraph{Internet-scale pretraining (Q4).}
We skip the pretraining stage and initialize the transformer with random weights, training directly on the per-task fine-tuning set with $\mathcal{L}_{\mathrm{ft}}$. This ablation achieves 67.5\% on Pick and Place.
\section{Conclusion}
\label{sec:conclusion}
In this paper, we propose \method, a framework for learning dexterous robotic manipulation policies from human demonstrations alone, taking a step toward human-to-robot transfer and data scaling in robot policy learning.
We tackle the embodiment gap between humans and robots, which has remained a major bottleneck for leveraging internet-scale human videos in dexterous manipulation.
Specifically, we propose a unified six-keypoint hand abstraction shared by the human demonstrator and the robot end-effector, which allows direct policy transfer without robot teleoperation.
To further compensate for the missing force modality inherent to point-only representations, we introduce a lightweight contact-point prediction mechanism that requires only minimal annotation.
Combined with internet-scale human-video pretraining on the VITRA corpus, our method achieves strong performance on a suite of real-world dexterous tasks with zero robot demonstrations at any stage of training.
Overall, our work highlights the effectiveness of point-based representations as a bridge between human video data and robotic action, and we believe this paradigm could inspire scalable approaches to dexterous robot learning in the future.

\paragraph{Limitations.}
Several limitations remain. First, although the keypoint representation loosens the coupling between human and robot hand actions, some predicted trajectories are still kinematically infeasible on the deployment robot and induce non-trivial IK error. Second, point-based observations carry no explicit force information; our fingertip contact annotation is a simple proxy, and richer signals such as tactile gloves would be a natural extension. Third, \method{} inherits the failure modes of the VLMs and other vision models it builds on, and is therefore tied to their continued improvement. Finally, point abstractions discard scene context that can matter in cluttered environments; hybrid representations that retain sparse visual context around keypoints are a promising direction.

\begin{ack}
This research was supported by the ``Advanced GPU Utilization Support Program'' funded by the Government of the Republic of Korea (Ministry of Science and ICT).
\end{ack}

\clearpage
{\small
\bibliography{ref}

@article{achiam2023gpt,
  title   = {{GPT-4} Technical Report},
  author  = {Achiam, Josh and Adler, Steven and Agarwal, Sandhini and Ahmad, Lama and Akkaya, Ilge and Aleman, Florencia Leoni and Almeida, Diogo and Altenschmidt, Janko and Altman, Sam and Anadkat, Shyamal and others},
  journal = {arXiv preprint arXiv:2303.08774},
  year    = {2023}
}

@article{touvron2023llama,
  title   = {{LLaMA}: Open and Efficient Foundation Language Models},
  author  = {Touvron, Hugo and Lavril, Thibaut and Izacard, Gautier and Martinet, Xavier and Lachaux, Marie-Anne and Lacroix, Timoth\'ee and Rozi\`ere, Baptiste and Goyal, Naman and Hambro, Eric and Azhar, Faisal and others},
  journal = {arXiv preprint arXiv:2302.13971},
  year    = {2023}
}

@article{team2023gemini,
  title   = {Gemini: A Family of Highly Capable Multimodal Models},
  author  = {{Gemini Team}},
  journal = {arXiv preprint arXiv:2312.11805},
  year    = {2023}
}

@article{ramesh2022dalle,
  title   = {Hierarchical Text-Conditional Image Generation with {CLIP} Latents},
  author  = {Ramesh, Aditya and Dhariwal, Prafulla and Nichol, Alex and Chu, Casey and Chen, Mark},
  journal = {arXiv preprint arXiv:2204.06125},
  year    = {2022}
}

@inproceedings{rombach2022stablediffusion,
  title     = {High-Resolution Image Synthesis with Latent Diffusion Models},
  author    = {Rombach, Robin and Blattmann, Andreas and Lorenz, Dominik and Esser, Patrick and Ommer, Bj\"orn},
  booktitle = {CVPR},
  year      = {2022}
}

@misc{brooks2024sora,
  title        = {Video Generation Models as World Simulators},
  author       = {Brooks, Tim and Peebles, Bill and Holmes, Connor and DePue, Will and Guo, Yufei and Jing, Li and Schnurr, David and Taylor, Joe and Luhman, Troy and Luhman, Eric and Ng, Clarence and Wang, Ricky and Ramesh, Aditya},
  year         = {2024},
  howpublished = {OpenAI Technical Report},
  url          = {https://openai.com/research/video-generation-models-as-world-simulators}
}

@inproceedings{yang2024cogvideox,
  title     = {{CogVideoX}: Text-to-Video Diffusion Models with An Expert Transformer},
  author    = {Yang, Zhuoyi and Teng, Jiayan and Zheng, Wendi and Ding, Ming and Huang, Shiyu and Xu, Jiazheng and Yang, Yuanming and Hong, Wenyi and Zhang, Xiaohan and Feng, Guanyu and Yin, Da and Gu, Xiaotao and Zhang, Yuxuan and Wang, Weihan and Cheng, Yean and Liu, Ting and Xu, Bin and Dong, Yuxiao and Tang, Jie},
  booktitle = {ICLR},
  year      = {2025}
}

@inproceedings{zhao2023aloha,
  title     = {Learning Fine-Grained Bimanual Manipulation with Low-Cost Hardware},
  author    = {Zhao, Tony Z. and Kumar, Vikash and Levine, Sergey and Finn, Chelsea},
  booktitle = {RSS},
  year      = {2023}
}

@inproceedings{khazatsky2024droid,
  title     = {{DROID}: A Large-Scale In-the-Wild Robot Manipulation Dataset},
  author    = {Khazatsky, Alexander and Pertsch, Karl and Nair, Suraj and Balakrishna, Ashwin and Dasari, Sudeep and Karamcheti, Siddharth and Nasiriany, Soroush and Srirama, Mohan Kumar and Chen, Lawrence Yunliang and Ellis, Kirsty and others},
  booktitle = {RSS},
  year      = {2024}
}

@inproceedings{o2023openx,
  title     = {Open {X-Embodiment}: Robotic Learning Datasets and {RT-X} Models},
  author    = {{Open X-Embodiment Collaboration}},
  booktitle = {ICRA},
  year      = {2024}
}

@inproceedings{grauman2022ego4d,
  title     = {{Ego4D}: Around the World in 3,000 Hours of Egocentric Video},
  author    = {Grauman, Kristen and others},
  booktitle = {CVPR},
  year      = {2022}
}

@inproceedings{grauman2024egoexo4d,
  title     = {{Ego-Exo4D}: Understanding Skilled Human Activity from First- and Third-Person Perspectives},
  author    = {Grauman, Kristen and others},
  booktitle = {CVPR},
  year      = {2024}
}

@inproceedings{goyal2017something,
  title     = {The ``Something Something'' Video Database for Learning and Evaluating Visual Common Sense},
  author    = {Goyal, Raghav and others},
  booktitle = {ICCV},
  year      = {2017}
}

@inproceedings{damen2018scaling,
  title     = {Scaling Egocentric Vision: The {EPIC-KITCHENS} Dataset},
  author    = {Damen, Dima and others},
  booktitle = {ECCV},
  year      = {2018}
}

@inproceedings{nair2023r3m,
  title     = {{R3M}: A Universal Visual Representation for Robot Manipulation},
  author    = {Nair, Suraj and Rajeswaran, Aravind and Kumar, Vikash and Finn, Chelsea and Gupta, Abhinav},
  booktitle = {CoRL},
  year      = {2022}
}

@inproceedings{ma2023vip,
  title     = {{VIP}: Towards Universal Visual Reward and Representation via Value-Implicit Pre-Training},
  author    = {Ma, Yecheng Jason and others},
  booktitle = {ICLR},
  year      = {2023}
}

@inproceedings{karamcheti2023language,
  title     = {Language-Driven Representation Learning for Robotics},
  author    = {Karamcheti, Siddharth and others},
  booktitle = {RSS},
  year      = {2023}
}

@article{majumdar2023vc1,
  title   = {Where are we in the search for an Artificial Visual Cortex for Embodied Intelligence?},
  author  = {Majumdar, Arjun and others},
  journal = {arXiv preprint arXiv:2303.18240},
  year    = {2023}
}

@inproceedings{bahl2023affordance,
  title     = {Affordances from Human Videos as a Versatile Representation for Robotics},
  author    = {Bahl, Shikhar and Mendonca, Russell and Chen, Lili and Jain, Unnat and Pathak, Deepak},
  booktitle = {CVPR},
  year      = {2023}
}

@article{bharadhwaj2024track2act,
  title   = {{Track2Act}: Predicting Point Tracks from Internet Videos enables Generalizable Robot Manipulation},
  author  = {Bharadhwaj, Homanga and Mottaghi, Roozbeh and Gupta, Abhinav and Tulsiani, Shubham},
  journal = {arXiv preprint arXiv:2405.01527},
  year    = {2024}
}

@article{bharadhwaj2024gen2act,
  title   = {{Gen2Act}: Human Video Generation in Novel Scenarios enables Generalizable Robot Manipulation},
  author  = {Bharadhwaj, Homanga and others},
  journal = {arXiv preprint arXiv:2409.16283},
  year    = {2024}
}

@inproceedings{wen2023motiontracks,
  title     = {Any-point Trajectory Modeling for Policy Learning},
  author    = {Wen, Chuan and Lin, Xingyu and So, John and Chen, Kai and Dou, Qi and Gao, Yang and Abbeel, Pieter},
  booktitle = {Robotics: Science and Systems (RSS)},
  year      = {2024}
}

@inproceedings{shaw2023videodex,
  title     = {{VideoDex}: Learning Dexterity from Internet Videos},
  author    = {Shaw, Kenneth and Bahl, Shikhar and Pathak, Deepak},
  booktitle = {CoRL},
  year      = {2023}
}

@inproceedings{mandikal2022dexvip,
  title     = {{DexVIP}: Learning Dexterous Grasping with Human Hand Pose Priors from Video},
  author    = {Mandikal, Priyanka and Grauman, Kristen},
  booktitle = {Conference on Robot Learning (CoRL)},
  year      = {2022}
}

@article{lepert2025phantom,
  title   = {{Phantom}: Training Robots Without Robots Using Only Human Videos},
  author  = {Lepert, Matthieu and others},
  journal = {arXiv preprint arXiv:2503.00779},
  year    = {2025}
}

@article{intelligence2025pi05,
  title   = {{$\pi_{0.5}$}: A Vision-Language-Action Model with Open-World Generalization},
  author  = {{Physical Intelligence}},
  journal = {arXiv preprint arXiv:2504.16054},
  year    = {2025}
}

@article{bjorck2025gr00t,
  title   = {{GR00T N1}: An Open Foundation Model for Generalist Humanoid Robots},
  author  = {Bjorck, Johan and others},
  journal = {arXiv preprint arXiv:2503.14734},
  year    = {2025}
}

@inproceedings{manuelli2019kpam,
  title     = {{kPAM}: KeyPoint Affordances for Category-Level Robotic Manipulation},
  author    = {Manuelli, Lucas and Gao, Wei and Florence, Peter and Tedrake, Russ},
  booktitle = {ISRR},
  year      = {2019}
}

@inproceedings{simeonov2022ndf,
  title     = {Neural Descriptor Fields: {SE(3)}-Equivariant Object Representations for Manipulation},
  author    = {Simeonov, Anthony and Du, Yilun and Tagliasacchi, Andrea and Tenenbaum, Joshua B. and Rodriguez, Alberto and Agrawal, Pulkit and Sitzmann, Vincent},
  booktitle = {ICRA},
  year      = {2022}
}

@inproceedings{huang2024rekep,
  title     = {{ReKep}: Spatio-Temporal Reasoning of Relational Keypoint Constraints for Robotic Manipulation},
  author    = {Huang, Wenlong and Wang, Chen and Li, Yunzhu and Zhang, Ruohan and Fei-Fei, Li},
  booktitle = {CoRL},
  year      = {2024}
}

@inproceedings{liu2024moka,
  title     = {{MOKA}: Open-World Robotic Manipulation through Mark-Based Visual Prompting},
  author    = {Liu, Fangchen and Fang, Kuan and Abbeel, Pieter and Levine, Sergey},
  booktitle = {RSS},
  year      = {2024}
}

@inproceedings{yuan2024robopoint,
  title     = {{RoboPoint}: A Vision-Language Model for Spatial Affordance Prediction for Robotics},
  author    = {Yuan, Wentao and Duan, Jiafei and Blukis, Valts and Pumacay, Wilbert and Krishna, Ranjay and Murali, Adithyavairavan and Mousavian, Arsalan and Fox, Dieter},
  booktitle = {CoRL},
  year      = {2024}
}

@article{levy2024p3po,
  title   = {{P3-PO}: Prescriptive Point Priors for Visuo-Spatial Generalization of Robot Policies},
  author  = {Levy, Mara and Haldar, Siddhant and Pinto, Lerrel and Shrivastava, Abhinav},
  journal = {arXiv preprint arXiv:2412.06784},
  year    = {2024}
}

@inproceedings{pointpolicy,
  title     = {Point Policy: Unifying Observations and Actions with Key Points for Robot Manipulation},
  author    = {Haldar, Siddhant and Pinto, Lerrel},
  booktitle = {Conference on Robot Learning (CoRL)},
  year      = {2025}
}

@article{pointbridge,
  title   = {Point Bridge: {3D} Representations for Cross Domain Policy Learning},
  author  = {Haldar, Siddhant and Johannsmeier, Lars and Pinto, Lerrel and Gupta, Abhishek and Fox, Dieter and Narang, Yashraj and Mandlekar, Ajay},
  journal = {arXiv preprint arXiv:2601.16212},
  year    = {2026}
}

@inproceedings{vitra2025,
  title     = {Scalable Vision-Language-Action Model Pretraining for Robotic Manipulation with Real-Life Human Activity Videos},
  author    = {Li, Qixiu and Deng, Yu and Liang, Yaobo and Luo, Lin and Zhou, Lei and Yao, Chengtang and Zeng, Lingqi and Feng, Zhiyuan and Liang, Huizhi and Xu, Sicheng and Zhang, Yizhong and Chen, Xi and Chen, Hao and Sun, Lily and Chen, Dong and Yang, Jiaolong and Guo, Baining},
  booktitle = {IEEE International Conference on Robotics and Automation (ICRA)},
  year      = {2026}
}

@inproceedings{qi2017pointnet,
  title     = {{PointNet}: Deep Learning on Point Sets for {3D} Classification and Segmentation},
  author    = {Qi, Charles R. and Su, Hao and Mo, Kaichun and Guibas, Leonidas J.},
  booktitle = {CVPR},
  year      = {2017}
}

@inproceedings{reimers2019sentencebert,
  title     = {{Sentence-BERT}: Sentence Embeddings using {Siamese BERT}-Networks},
  author    = {Reimers, Nils and Gurevych, Iryna},
  booktitle = {EMNLP},
  year      = {2019}
}

@article{bai2025qwen3vl,
  title   = {{Qwen3-VL} Technical Report},
  author  = {Bai, Shuai and others},
  journal = {arXiv preprint arXiv:2511.21631},
  year    = {2025}
}

@article{meta2025sam3,
  title   = {{SAM} 3: Segment Anything with Concepts},
  author  = {Carion, Nicolas and Gustafson, Laura and Hu, Yuan-Ting and Debnath, Shoubhik and Hu, Ronghang and Suris, Didac and Ryali, Chaitanya and Alwala, Kalyan Vasudev and Khedr, Haitham and others},
  journal = {arXiv preprint arXiv:2511.16719},
  year    = {2025}
}

@article{depthanything3,
  title   = {{Depth Anything} 3: Recovering the Visual Space from Any Views},
  author  = {Lin, Haotong and Chen, Sili and Liew, Jun Hao and Chen, Donny Y. and Li, Zhenyu and Shi, Guang and Feng, Jiashi and Kang, Bingyi},
  journal = {arXiv preprint arXiv:2511.10647},
  year    = {2025}
}

@inproceedings{hawor2024,
  title     = {{HaWoR}: World-Space Hand Motion Reconstruction from Egocentric Videos},
  author    = {Zhang, Jinglei and Deng, Jiankang and Ma, Chao and Potamias, Rolandos Alexandros},
  booktitle = {Proceedings of the IEEE/CVF Conference on Computer Vision and Pattern Recognition (CVPR)},
  year      = {2025}
}

@article{romero2022embodied,
  title={Embodied hands: Modeling and capturing hands and bodies together},
  author={Romero, Javier and Tzionas, Dimitrios and Black, Michael J},
  journal={arXiv preprint arXiv:2201.02610},
  year={2022}
}

@inproceedings{pavlakos2024reconstructing,
  title={Reconstructing hands in 3d with transformers},
  author={Pavlakos, Georgios and Shan, Dandan and Radosavovic, Ilija and Kanazawa, Angjoo and Fouhey, David and Malik, Jitendra},
  booktitle={Proceedings of the IEEE/CVF Conference on Computer Vision and Pattern Recognition},
  pages={9826--9836},
  year={2024}
}

@inproceedings{
  li2025reinforcement,
  title={Reinforcement Learning with Action Chunking},
  author={Qiyang Li and Zhiyuan Zhou and Sergey Levine},
  booktitle={The Thirty-ninth Annual Conference on Neural Information Processing Systems},
  year={2025},
}

@misc{ankile2025residualoffpolicyrlfinetuning,
        title={Residual Off-Policy RL for Finetuning Behavior Cloning Policies}, 
        author={Lars Ankile and Zhenyu Jiang and Rocky Duan and Guanya Shi and Pieter Abbeel and Anusha Nagabandi},
        year={2025},
        eprint={2509.19301},
        archivePrefix={arXiv},
        primaryClass={cs.RO},
  }

@inproceedings{
chen2021randomized,
title={Randomized Ensembled Double Q-Learning: Learning Fast Without a Model},
author={Xinyue Chen and Che Wang and Zijian Zhou and Keith W. Ross},
booktitle={International Conference on Learning Representations},
year={2021},
}
\bibliographystyle{unsrtnat}
}

\clearpage
\appendix
\crefalias{section}{appendix}
\section{Training Details}
\label{app:training}

\paragraph{Pretraining.}
We pretrain the autoregressive transformer on the VITRA corpus ($\sim$1M egocentric episodes) for $100$k optimizer steps using AdamW with learning rate $10^{-4}$, weight decay $10^{-4}$, and a global batch size of $256$. Training uses bf16 mixed precision (forward and loss in bf16, optimizer step in fp32), gradient clipping at $\lVert g\rVert\le 1$, and a linear warmup schedule (\texttt{LinearLR} with \texttt{start\_factor}$=10^{-2}$ over the first $1$k steps, then constant). Each forward pass produces an action chunk of length $Q=16$ with teacher-forced targets. Pretraining runs on $\text{1}\times$ NVIDIA A100 (80\,GB) GPU and takes approximately 36 GPU-hours.

\paragraph{Fine-tuning.}
We fine-tune from the pretrained $100$k checkpoint, keeping the optimizer, weight decay, learning rate, gradient clipping, and mixed-precision settings of pretraining. For Pick and Place, we fine-tune on $500$ human demonstrations spanning $5$ objects (bottle, box, gray tennis ball, white stuffed bear, gray towel; $100$ demonstrations per object) for $400$k steps with batch size $128$. For each Manipulation \& Tool Use task (open, spray, brush) we fine-tune separately on the $100$ demonstrations of that task for the same number of steps; we use batch size $64$ for Tool runs. The contact loss weight is $\lambda=1$ throughout, and the contact-head gradient is detached from the GPT backbone so contact supervision only updates the contact head. Fine-tuning takes approximately 4 hours per task on a single A100 GPU. The action loss is applied to the six robot keypoints only; object-point tokens carry observations and receive no supervision.

\section{Baseline Details}
\label{app:baselines}

\paragraph{Point Policy.}
We follow the original Point Policy architecture~\citep{pointpolicy}: a non-causal (bidirectional) transformer that takes the same tokenized observation as \method{} (language, object points, and hand keypoints) and predicts the full $H$-step action chunk for every robot keypoint in a single forward pass via an MLP action head. Following the original formulation, all observation tokens and the trailing robot-keypoint tokens attend to one another bidirectionally; there is no causal mask and no autoregressive rollout across the chunk dimension. We replace the original two-keypoint gripper representation with our six-keypoint hand abstraction (wrist plus five fingertips) so that the action space matches \method{}'s. The model is trained from scratch on the same per-task human demonstrations as \method{}, with no VITRA pretraining and no contact channel; the only supervision is the keypoint regression loss $\mathcal{L}_{\mathrm{act}}$. At deployment, predicted keypoints are mapped to joint targets via the same damped-least-squares IK solver as \method{} (\cref{subsec:inference}), but without the contact-driven closing offset. Holding the observation, action space, and IK fixed, this baseline isolates the combined effect of our three contributions: internet-scale pretraining, autoregressive decoding across the chunk, and the contact prediction head.

\paragraph{VITRA.}
We use the VITRA model~\citep{vitra2025}, which is pretrained on the same internet-scale egocentric video corpus that \method{} uses for pretraining. Since VITRA's action space is defined in robot joint space and our setting contains no robot demonstrations, we construct fine-tuning supervision as follows: for each human demonstration, we run our hand-point extraction pipeline to obtain the six-keypoint hand trajectory, then convert it to a robot joint trajectory using the same damped-least-squares IK solver used at \method{}'s deployment (\cref{subsec:inference}). The resulting joint trajectories serve as VITRA's fine-tuning targets. This adaptation ensures that VITRA receives exactly the same information content as \method{}. At deployment, VITRA's predicted joint angles are sent directly to the controller, bypassing the IK stage that \method{} uses to map keypoint targets to joint commands.

\section{Experiment Setup Details}
\label{app:setup}

\begin{figure}[t]
    \centering
    \includegraphics[width=\linewidth]{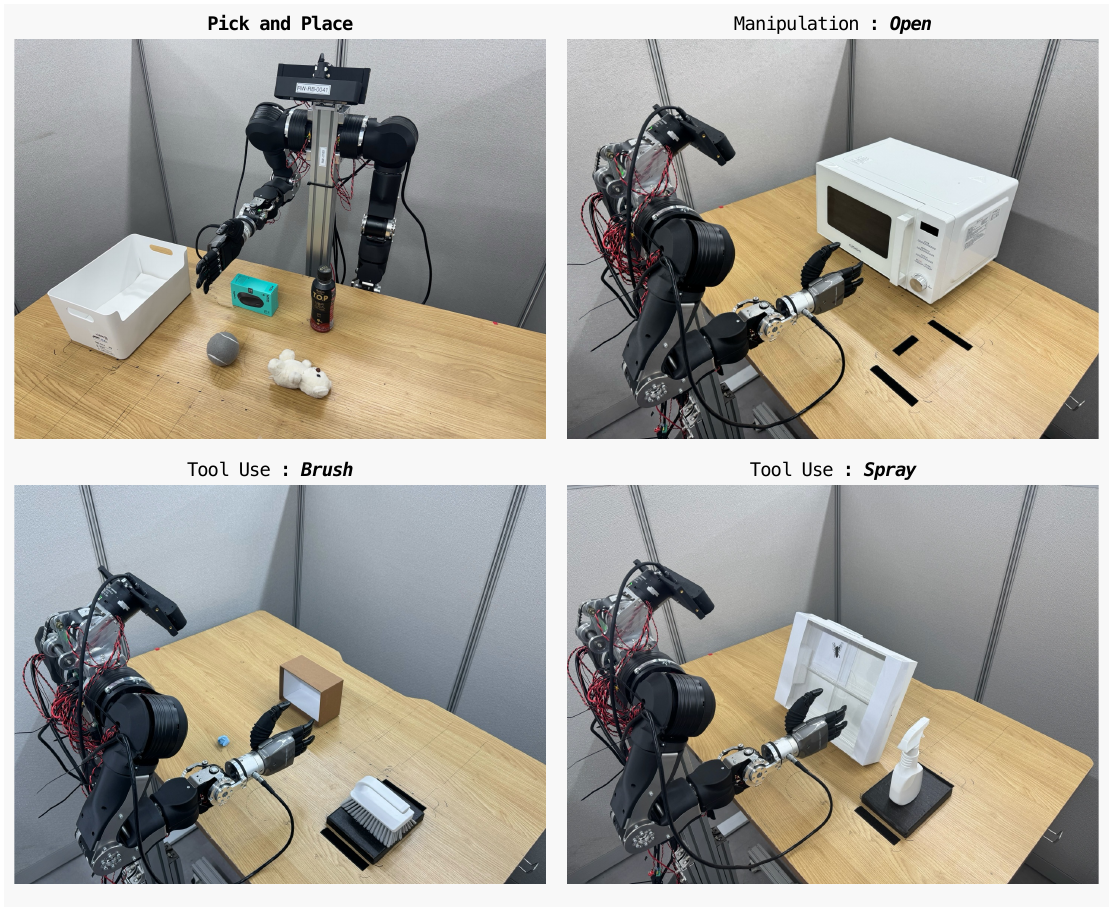}
    \caption{\textbf{Real-world task setup.} We evaluate \method{} on four task categories: \emph{Pick and Place} (top-left) with five objects placed on a $2\times2$ grid and a fixed target container; \emph{Open} (top-right), where the robot opens a microwave door from a randomized initial pose; \emph{Brush} (bottom-left), where the robot grasps a hand brush and sweeps debris to a target location; and \emph{Spray} (bottom-right), where the robot grasps a spray bottle and depresses the trigger toward a target on a window.}
    \label{fig:task-setup}
\end{figure}

\begin{figure}[t]
    \centering
    \includegraphics[width=\linewidth]{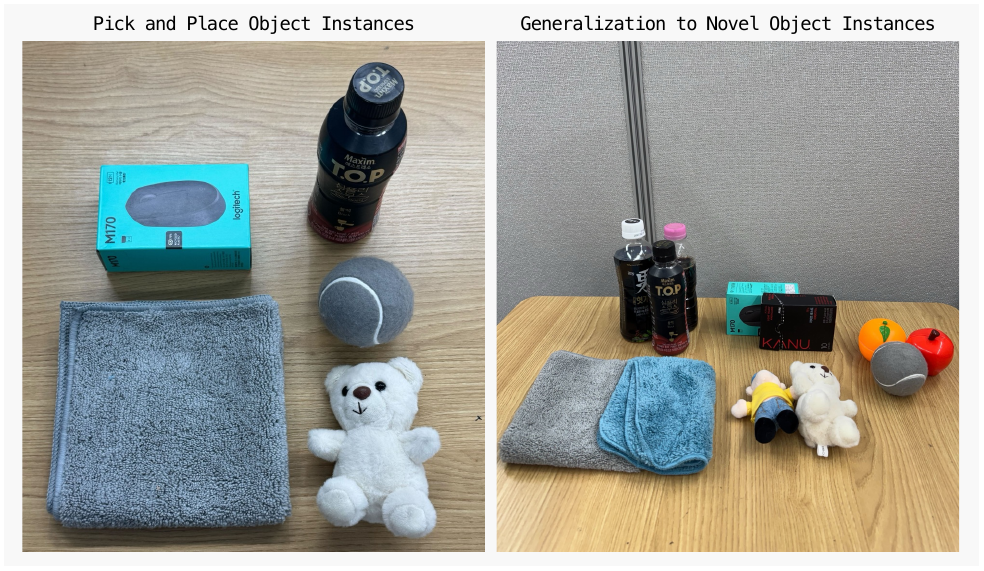}
    \caption{\textbf{Pick and Place object instances.} Left: the five training objects spanning four shape categories --- bottle (cylindrical), box (rectangular), tennis ball (spherical), towel and teddy bear (deformable/irregular). Right: novel object instances used in the generalization evaluation (\cref{subsec:exp_generalization}), which share the broad shape category of their training counterparts but differ in size, color, and surface texture.}
    \label{fig:pnp-objects}
\end{figure}

\paragraph{Pick and Place.}
The four candidate object positions are arranged in a $2\times2$ grid on the table. The target container is placed at a fixed location to the right of the grid center. In each trial, a single object is placed at one of the four grid positions. We use five objects: bottle (cylindrical), tennis ball (spherical), box (rectangular), towel (deformable/irregular), and teddy bear (deformable/irregular). The training and novel object instances used in our evaluation are shown in \cref{fig:pnp-objects}. In the multi-object evaluation (\cref{subsec:exp_generalization}), all four positions are occupied simultaneously with different objects. For the novel object evaluation, we additionally introduce one or two previously unseen objects with similar size, shape, and appearance to those used during training.

\paragraph{Open Microwave.}
The microwave is initialized as fully closed, placed at one of four randomized locations on the table.

\paragraph{Brush.}
The brush is placed at a fixed initial position on the table, and a small piece of debris is placed at one of four locations on a $2\times2$ grid. The robot must grasp the hand brush, approach the debris, and sweep it to a fixed destination on the table.

\paragraph{Spray.}
The spray bottle is placed at a fixed initial position on the table. The robot must grasp the bottle body with a stable grip and depress the trigger using its thumb and index finger. The spray target is a $2\times2$ grid of regions on a window, where a bug is placed at one of the four locations; the objective is to orient the bottle toward the bug and actuate the trigger so that the spray is directed at the correct position.

\section{Success Criteria}
\label{app:success}

We define task success as follows. Success judgments are made by a human evaluator observing the trial in real time.

\begin{itemize}[leftmargin=1.5em,itemsep=2pt]
  \item \textbf{Pick and Place.} The robot grasps the target object, lifts it above the table surface, transports it to the container, and releases it inside. A trial is successful if the object comes to rest inside the container. Considering recovery, we set 1 minute as the timeout.
  \item \textbf{Open.} The robot grasps the microwave door handle and pulls it open. A trial is successful if the door latch disengages and the door opens past the closed position. Considering recovery, we set 3 minutes as the timeout.
  \item \textbf{Brush.} The robot grasps the brush, secures it properly in its end-effector, and sweeps debris toward the designated target position. A trial is successful if the robot maintains a stable grasp on the brush throughout the motion and transports the debris into the target position. Considering recovery, we set 3 minutes as the timeout.
  \item \textbf{Spray.} The robot grasps the spray bottle, aligns its nozzle toward a designated target position, and depresses the trigger. A trial is successful if the bottle is oriented toward the target and the trigger is actuated to release spray in that direction. Considering recovery, we set 3 minutes as the timeout.
\end{itemize}

\section{Potential Societal Impact}
\label{app:societal-impact}
While the keypoint-based data extraction and the policy design of~\method~can be beneficial for various robotic applications, such as object manipulation, tool use, and humanoids, the emergence of unexpected behavior within~\method~can lead to misrepresentations of the safe action data. 
For those applications that require extremely accurate models for safety-related judgments, such as policy learning for assistive humanoid robots for disabled persons or children, the unexpected behaviors must be carefully managed. 
To ensure the reliability of systems using point-based action predictions,
we recommend that one conduct thorough investigations and implement robust mitigation strategies to minimize potential risks, thereby increasing the overall safety and effectiveness of these applications.

\section{Scale-Consistent HaWoR}
\label{app:hand-tracker}

\begin{figure}[t]
\centering
\begin{subfigure}{0.48\textwidth}
\includegraphics[width=\textwidth]{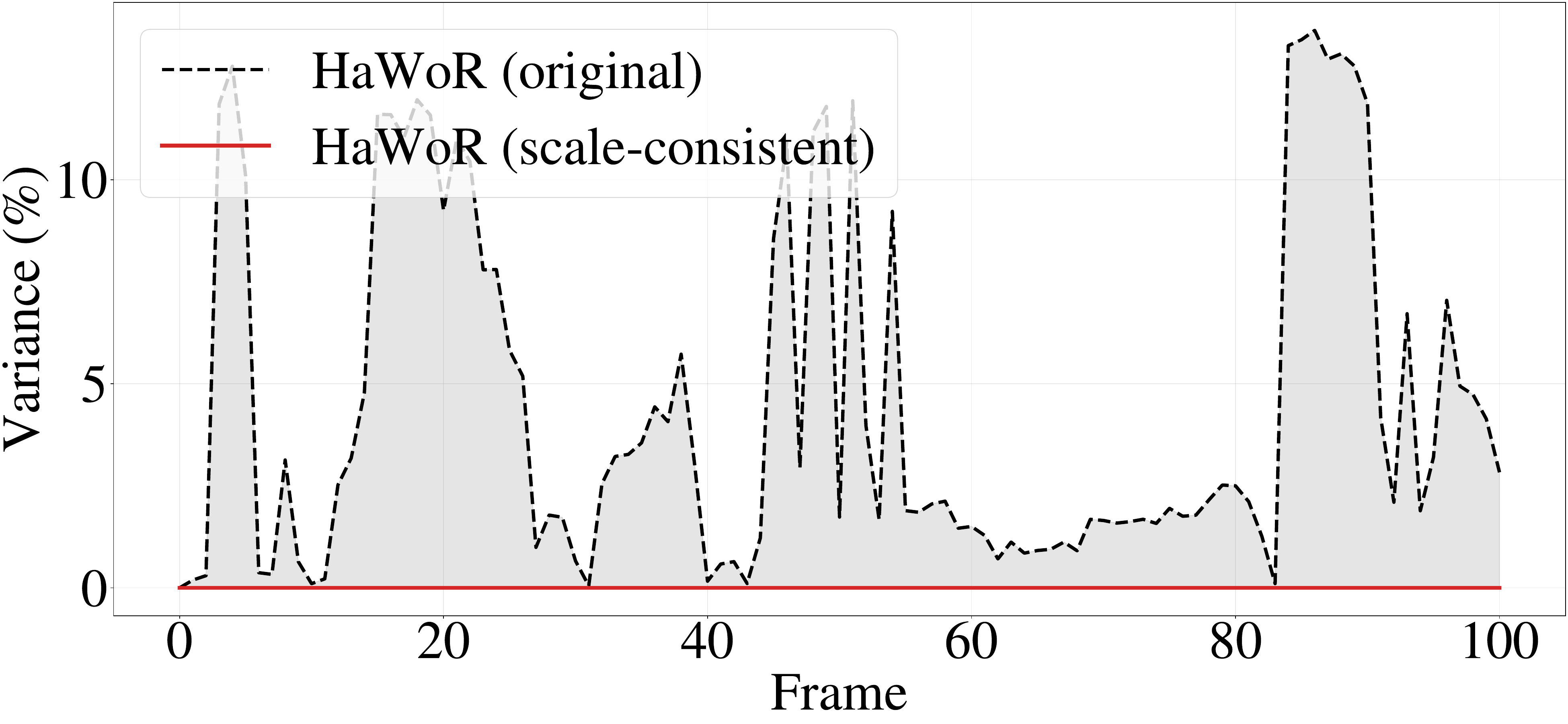}
\caption{Hand scale variance}
\label{fig:hawor1}
\end{subfigure}
\begin{subfigure}{0.48\textwidth}
\includegraphics[width=\textwidth]{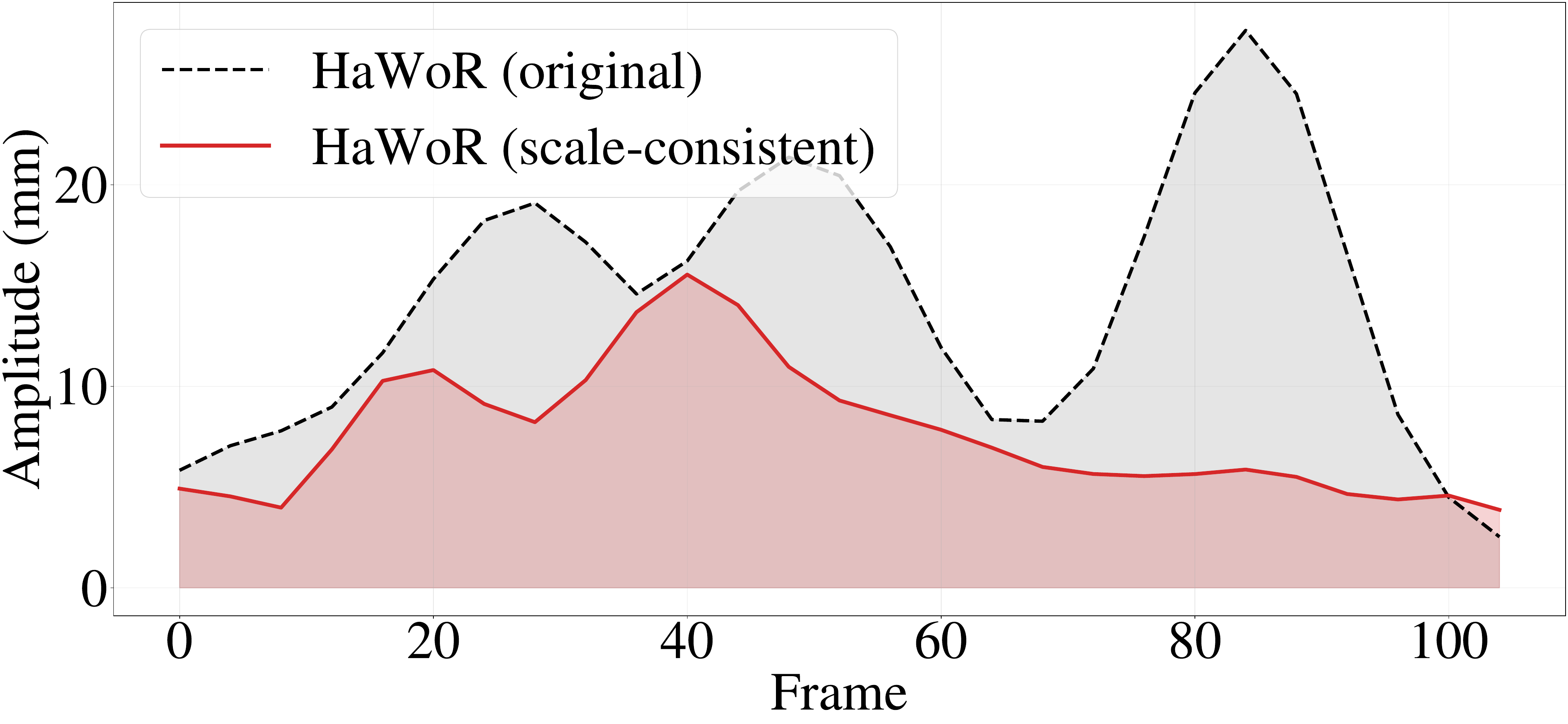}
\caption{Noise amplitude}
\label{fig:hawor2}
\end{subfigure}
\caption{\textbf{Comparison between original and scale-consistent HaWoR.} We compare inference results on a single egocentric video with each model, and visualize (a) hand scale variance, and (b) noise amplitude.
Indeed, the original HaWoR shows more than 10\% scale inconsistency even within a single egocentric video, while also showing higher noise amplitude along the depth axis.
}
\label{fig:hawor}
\end{figure}

With the given egocentric video $\mathbf{V}\in\mathbb{R}^{T\times H\times W \times 3}$, the hand motion estimation network $\mathcal{M}$ of \texttt{HaWoR}~\citep{hawor2024} yields MANO \cite{romero2022embodied} shape $\{\beta^i_t \in \mathbb{R}^{10}\}_{t=0}^T$ and pose parameters $\{\Theta \in \mathbb{R}^{15 \times 3}\}_{t=0}^T$ along with the 6d pose of the wrist $\hat{T} = (\hat{R}\in\mathbb{R}^{3}, \hat{t}\in\mathbb{R}^{3})$ \textit{per frame}.
However, as shown in \cref{fig:hawor1}, even within a single egocentric video, the hand scale varies notably, \ie $>10cm$.
In other words, the model inconsistently infers that larger hands are farther away and smaller hands are closer than the ground truth.
To verify this, in \cref{fig:hawor2}, we visualize the depth noise amplitude, \ie $> 30\text{Hz}$ signal which is definitely noise from the human action within a single sequence.
Indeed, the original \texttt{HaWoR} contains notably higher noise along the depth axis, which hinders the model from learning the clean trajectory of the human wrist.
To this end, we propose a modified hand motion estimation network $\mathcal{M}^{\prime}$ which utilizes the shape parameter $\beta$ as an input instead of the output:
\begin{equation}
\mathcal{M}^{\prime}(\beta, V) := (\Theta, \hat{T})
\label{eq:sc-hawor}
\end{equation}
In the training phase, we use the ground truth $\beta$ as an input of $\mathcal{M}^{\prime}$.
Since there is no ground truth $\beta$ during the inference stage, we use the mean value of shape parameters among the frame-wise inference results, \ie $\overline\beta$, obtained by HaMeR~\citep{pavlakos2024reconstructing}.

\section{Auxiliary Residual RL Adaptation for the Dexterous Point Policy}
\label{app:dpp-residual-rl}

\subsection{Motivation and Scope}
\label{app:dpp-rl-motivation}

\paragraph{Purpose.}
We show an \emph{auxiliary experiment} that is separate from the main training pipeline of the paper.
The dexterous point policy in the main paper is trained from human videos only and does not use real-robot demonstrations, robot teleoperation, or robot-specific fine-tuning.
Here, we conduct a separate simulation study to examine whether this trained policy can serve as a base policy for residual reinforcement learning.
The goal of this experiment is not to propose a new RL algorithm, but to examine whether the trained policy provides a useful action prior for downstream residual RL adaptation.

\paragraph{Question.}
This auxiliary experiment is designed to answer the following question:
\begin{itemize}[leftmargin=1.5em,itemsep=2pt]
    \item \textbf{AQ1.} Can a frozen dexterous point policy serve as a base policy for residual reinforcement learning?
\end{itemize}

We answer this question by freezing the trained dexterous point policy and using it as the base policy in the residual RL setup: its parameters are never updated, and RL only learns residual corrections on top of its predicted keypoint action chunks.
The base policy predicts a chunk of future keypoint actions, the residual policy refines this chunk, and the corrected actions are then executed sequentially through the same IK solver and low-level controller.
This experiment therefore evaluates whether the dexterous point policy can support downstream RL adaptation, separately from the main human-video-only training claim.

\subsection{Task and Residual RL Formulation}
\label{app:dpp-rl-formulation}

\begin{figure}[t]
    \centering
    \includegraphics[width=0.5\linewidth]{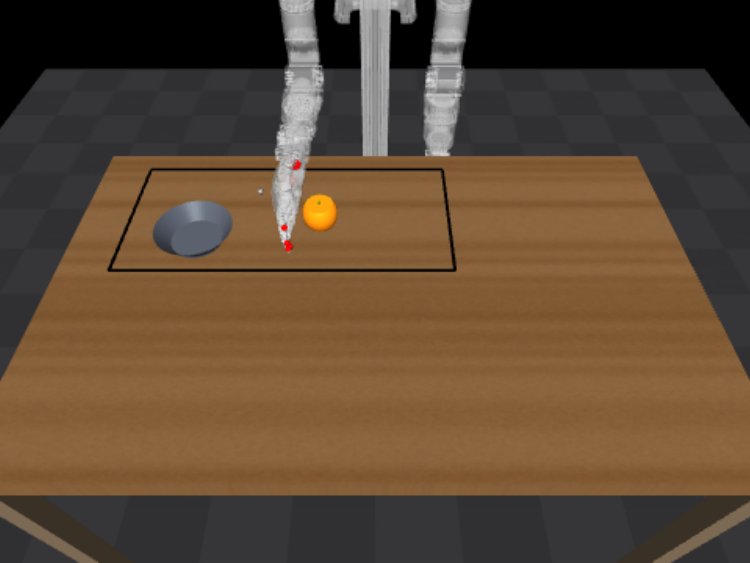}
    \caption{\textbf{Simulation setup for auxiliary residual RL.}}
    \label{fig:dpp-rl-sim-setup}
\end{figure}

\paragraph{Task, simulator, and initial states.}
The experiment is conducted in a MuJoCo dexterous-hand simulator configured to approximate the real-robot setup.
We use a spherical-object-to-bowl manipulation task, where the dexterous hand must move a spherical object from an initial table-top position into a bowl.
The object is initialized from one of four predefined anchor cells selected from a per-position base-policy success-rate sweep.
Each anchor cell corresponds to a $20\mathrm{mm}\times20\mathrm{mm}$ region.
During training, the anchor cell is sampled for each episode.
For evaluation, we use an anchor-balanced protocol over the same four anchor cells.
The exact anchor cells are listed in \cref{tab:dpp-rl-anchor-cells}, and \cref{tab:dpp-rl-formulation} summarizes the simulator and action-space settings.
\cref{fig:dpp-rl-sim-setup} shows the simulator layout used for this auxiliary experiment.

\paragraph{Reward.}
Each episode terminates when task success is achieved or when the maximum horizon is reached.
Success is defined by the simulator condition that the object is placed in the bowl.
We use a sparse binary success reward:
\begin{equation}
r_t =
\begin{cases}
1, & \text{if the simulator reports task success at step } t, \\
0, & \text{otherwise}.
\end{cases}
\label{eq:dpp-rl-sparse-reward}
\end{equation}

\paragraph{Method overview.}
The method combines residual action refinement with chunk-level value learning.
At each rollout step, the base policy predicts a chunk of future 18D keypoint actions.
The residual policy predicts a bounded correction chunk in the same normalized action space.
The corrected chunk is obtained by adding the residual chunk to the base chunk and clipping the result to the valid normalized action range.
The corrected actions inside the chunk are then executed sequentially using the same IK solver and low-level controller as the base policy.
The base policy is called again only after the current chunk has been executed.
During training, the critic evaluates the corrected action chunk rather than a single-step action, so residual RL refines the base policy at the chunk level.

\paragraph{Relation to ResFiT and Q-chunking.}
This experiment is not a direct reproduction of either ResFiT or Q-chunking.
From ResFiT~\citep{ankile2025residualoffpolicyrlfinetuning}, we borrow the structure of freezing a base policy and refining its action with a residual policy.
We adapt this idea to the dexterous point policy by using point encoders, proprioception, past contact, and base action chunks as inputs.
From Q-chunking~\citep{li2025reinforcement}, we borrow the temporally extended action space and the chunk-level critic target.
In our implementation, the base action chunk provides the prior, while the bounded residual scale and residual L2 penalty keep the learned correction close to the base behavior.

\paragraph{State.}
Residual RL uses the same geometric state abstraction as the base policy.
The critic does not directly consume visual appearance; instead, it uses object-wise 3D point clouds, robot hand keypoints, and proprioceptive features.
We denote the residual state at time $t$ as
\begin{equation}
s_t =
\left(
P^{\mathrm{hand}}_t,
P^{\mathrm{obj}}_t,
P^{\mathrm{goal}}_t,
q^{\mathrm{prop}}_t,
c^{\mathrm{past}}_t
\right).
\label{eq:dpp-rl-state}
\end{equation}

Here, $P^{\mathrm{hand}}_t \in \mathbb{R}^{6\times3}$ denotes the right-hand keypoints consisting of the wrist and five fingertips, $P^{\mathrm{obj}}_t \in \mathbb{R}^{128\times3}$ denotes the manipulated-object point cloud, and $P^{\mathrm{goal}}_t \in \mathbb{R}^{128\times3}$ denotes the bowl point cloud.
$q^{\mathrm{prop}}_t \in \mathbb{R}^{9}$ is a standardized wrist/proprioceptive feature, and $c^{\mathrm{past}}_t \in \mathbb{R}^{5}$ is the past-contact vector used as a contact-feedback input to the base policy.

The normalized base action chunk $\bar{a}^{b}_{t:t+H}$ is not treated as part of the environment state.
Instead, it is treated as a base-policy prior action chunk and is provided as an additional conditioning input to the residual policy and the critic.

The hand keypoints are passed through an MLP encoder, while the object and goal point sets are each passed through a PointNet-style encoder.
The resulting features are concatenated and combined with proprioception and past contact to form the state representation.
For residual learning, this state representation is further conditioned on the normalized base action chunk.
This provides a geometry-centric state abstraction that pairs naturally with a base-policy prior under sparse rewards.

\paragraph{Action parameterization.}
We use a keypoint action space in which each single-step action specifies the 3D target positions of six hand keypoints, corresponding to the wrist and five fingertips:
\begin{equation}
a_t
=
\left[
p^{(1)}_t,
p^{(2)}_t,
\ldots,
p^{(6)}_t
\right]
\in \mathbb{R}^{d_a},
\qquad
p^{(j)}_t \in \mathbb{R}^{3},
\quad
d_a = 6 \times 3.
\label{eq:dpp-rl-kp-action}
\end{equation}

The base policy predicts a horizon-$H$ keypoint action chunk:
\begin{equation}
a^{b}_{t:t+H}
=
\left[
a^{b}_t,
a^{b}_{t+1},
\ldots,
a^{b}_{t+H-1}
\right]
\in \mathbb{R}^{H\times d_a}.
\label{eq:dpp-rl-b-chunk}
\end{equation}

Residual addition is performed in a normalized keypoint action space.
Let $g(\cdot)$ denote a fixed action scaler and $g^{-1}(\cdot)$ its inverse.
The normalized base chunk is
\begin{equation}
\bar{a}^{b}_{t:t+H}
=
g\left(a^{b}_{t:t+H}\right).
\label{eq:dpp-rl-normalized-base}
\end{equation}

The scaler $g$ is fit once from the executed keypoint actions stored as \texttt{action\_kp} in the successful offline rollouts, and is fixed thereafter.
The residual policy predicts a bounded residual chunk in the same normalized space:
\begin{equation}
\bar{a}^{r}_{t:t+H}
=
\pi_\theta
\left(
s_t,
\bar{a}^{b}_{t:t+H}
\right),
\qquad
\bar{a}^{r}_{t:t+H}
\in
[-\alpha_{\mathrm{actor}}, \alpha_{\mathrm{actor}}]^{H\times d_a}.
\label{eq:dpp-rl-residual}
\end{equation}

The corrected normalized chunk is obtained by adding the base chunk and the residual chunk, followed by clipping:
\begin{equation}
\bar{a}_{t:t+H}
=
\mathrm{clip}
\left(
\bar{a}^{b}_{t:t+H}
+
\bar{a}^{r}_{t:t+H},
-1,1
\right).
\label{eq:dpp-rl-normalized-action}
\end{equation}

Finally, the corrected chunk is mapped back to the physical keypoint action space:
\begin{equation}
a_{t:t+H}
=
g^{-1}
\left(
\bar{a}_{t:t+H}
\right).
\label{eq:dpp-rl-action}
\end{equation}

Each step of $a_{t:t+H}$ is executed sequentially through the same inverse kinematics solver, contact-bend logic, and PI controller used by the base policy.
The base policy is called again only after the current chunk has been executed, unless the episode terminates earlier.
Thus, residual RL does not learn low-level motor control; it only learns local corrections to the base-policy trajectory.

\subsection{Training}
\label{app:dpp-rl-training}

\paragraph{Buffer initialization.}
The offline replay buffer is initialized with successful HDF5 rollout episodes collected in simulation by deploying the dexterous point policy trained from real-world human videos.
These simulation rollouts are not used to train the dexterous point policy in the main paper.
They are used only as an auxiliary replay source to stabilize this simulation-only residual adaptation experiment.

The replay buffer stores frame-level transitions.
Consecutive frames are grouped into chunks only at training time to build the Q-chunking target.
This preserves all frame transitions while allowing the critic to learn a temporally extended value
$Q(s_t,\bar{a}^{b}_{t:t+H},\bar{a}_{t:t+H})$.
Each minibatch mixes offline replay and online replay according to the ratio in \cref{tab:dpp-rl-hparams}.

\paragraph{Chunk-level critic target.}
Following Q-chunking, the critic evaluates temporally extended action chunks rather than single-step actions.
For a sampled chunk, the critic target accumulates the observed rewards within the valid portion of the chunk and, if the chunk does not terminate the episode, bootstraps from the next corrected action chunk.
In our setting, the chunk is formed by adding a bounded residual chunk to the base policy's keypoint action chunk.

Because a sampled chunk may contain a terminal transition before reaching the full horizon $H$, transitions after the first terminal transition are masked out.
Let $m_{i,k}\in\{0,1\}$ indicate whether transition $i+k$ is valid.
A transition is valid if it occurs before or at the first terminal transition in the sampled chunk.
Let $\ell_i=\sum_{k=0}^{H-1}m_{i,k}$ be the valid chunk length.
The observed chunk return is
\begin{equation}
R_i
=
\sum_{k=0}^{H-1}
m_{i,k}\gamma^k r_{i+k}.
\label{eq:dpp-rl-chunk-return}
\end{equation}

Let $d_i\in\{0,1\}$ indicate whether a terminal transition appears within the sampled chunk.
For non-terminal chunks, we build the next corrected normalized chunk with the target residual policy:
\begin{equation}
\bar{a}'_{i+\ell_i:i+\ell_i+H}
=
\mathrm{clip}
\left(
\bar{a}^{b}_{i+\ell_i:i+\ell_i+H}
+
\pi_{\theta'}
\left(
s_{i+\ell_i},
\bar{a}^{b}_{i+\ell_i:i+\ell_i+H}
\right)
+
\epsilon',
-1,1
\right).
\label{eq:dpp-rl-target-action}
\end{equation}
Here $\epsilon'$ denotes the clipped noise used for TD3-style target smoothing.

Let $Q_{\phi'_j}$ denote the target network corresponding to critic head $Q_{\phi_j}$.
We then compute the conservative target value
\begin{equation}
\widehat{Q}_i
=
\min_{j\in\mathcal{J}_2}
Q_{\phi'_j}
\left(
s_{i+\ell_i},
\bar{a}^{b}_{i+\ell_i:i+\ell_i+H},
\bar{a}'_{i+\ell_i:i+\ell_i+H}
\right),
\label{eq:dpp-rl-target-q}
\end{equation}
where $\mathcal{J}_2$ is a randomly sampled two-head subset of the target critic ensemble, following a REDQ-style conservative ensemble target~\citep{chen2021randomized}.

The TD target for critic learning is
\begin{equation}
y_i
=
R_i
+
(1-d_i)
\gamma^{\ell_i}
\widehat{Q}_i.
\label{eq:dpp-rl-compact-target}
\end{equation}
When a terminal transition appears inside the sampled chunk, $d_i=1$ and the bootstrap term is removed.
The critic is trained to regress each ensemble head toward $y_i$:
\begin{equation}
\mathcal{L}_{Q}
=
\frac{1}{N_Q}
\sum_{j=1}^{N_Q}
\mathbb{E}
\left[
\left(
Q_{\phi_j}
\left(
s_i,
\bar{a}^{b}_{i:i+H},
\bar{a}_{i:i+H}
\right)
-
y_i
\right)^2
\right].
\label{eq:dpp-rl-critic-loss}
\end{equation}

The actor objective is to choose a residual chunk that increases the value of the full corrected chunk while keeping the residual magnitude small:
\begin{equation}
\mathcal{L}_{\pi}
=
-
\mathbb{E}
\left[
\frac{1}{N_Q}
\sum_{j=1}^{N_Q}
Q_{\phi_j}
\left(
s_i,
\bar{a}^{b}_{i:i+H},
\mathrm{clip}
\left(
\bar{a}^{b}_{i:i+H}
+
\pi_\theta(s_i,\bar{a}^{b}_{i:i+H}),
-1,1
\right)
\right)
\right]
+
\lambda_{\mathrm{res}}
\left\|
\pi_\theta(s_i,\bar{a}^{b}_{i:i+H})
\right\|_2^2.
\label{eq:dpp-rl-actor-loss}
\end{equation}

The actor update maximizes the mean Q value over all critic heads, while the target uses the two-head minimum for conservative backup.
The offline replay is used as a persistent replay source for critic stabilization, rather than as a separate behavior-cloning objective.
Each minibatch mixes offline replay and online replay according to the ratio in \cref{tab:dpp-rl-hparams}.
\cref{alg:dpp-rl-resfit-qchunk} summarizes the training loop, and the full implementation settings are listed at the end of this appendix.

\begin{algorithm}[t]
\caption{Residual TD3 with Q-chunking keypoint chunks}
\label{alg:dpp-rl-resfit-qchunk}
\begin{algorithmic}[1]
\State \textbf{Input:} base policy $\pi_b$ (frozen), offline replay $\mathcal{D}_{\mathrm{off}}$, chunk horizon $H$
\State \textbf{Initialize:} residual policy $\pi_\theta$, Q ensemble $Q_{\phi_1},\ldots,Q_{\phi_N}$, point encoder $f_\omega$, online replay $\mathcal{D}_{\mathrm{on}}$
\State \textbf{Initialize:} target networks $\theta'\leftarrow\theta$, $\phi'_j\leftarrow\phi_j$, $\omega'\leftarrow\omega$

\While{$|\mathcal{D}_{\mathrm{on}}| < N_{\mathrm{start}}$}
    \State Run base policy and normalize its output to obtain $\bar{a}^{b}_{t:t+H}=g(a^{b}_{t:t+H})$
    \State Sample warmup residual noise $\epsilon_{t:t+H}$ with the same shape as $\bar{a}^{b}_{t:t+H}$
    \State Execute $\bar{a}_{t:t+H}=\mathrm{clip}(\bar{a}^{b}_{t:t+H}+\epsilon_{t:t+H},-1,1)$
    \State Store resulting frame transitions in $\mathcal{D}_{\mathrm{on}}$
\EndWhile

\Repeat
    \State Run base policy and normalize its output to obtain $\bar{a}^{b}_{t:t+H}=g(a^{b}_{t:t+H})$
    \State Sample exploration noise $\epsilon_{\mathrm{expl}}$ with the same shape as $\bar{a}^{b}_{t:t+H}$
    \State Predict residual chunk $\bar{a}^{r}_{t:t+H}\leftarrow\pi_\theta(s_t,\bar{a}^{b}_{t:t+H})+\epsilon_{\mathrm{expl}}$
    \State Execute $\bar{a}_{t:t+H}=\mathrm{clip}(\bar{a}^{b}_{t:t+H}+\bar{a}^{r}_{t:t+H},-1,1)$
    \State Store resulting frame transitions in $\mathcal{D}_{\mathrm{on}}$

    \For{$u=1,\ldots,\mathrm{UTD}$}
        \State Sample batch $B$ from $\mathcal{D}_{\mathrm{off}}$ and $\mathcal{D}_{\mathrm{on}}$ according to the offline/online ratio
        \State Assemble $H$-step chunks from frame transitions in $B$
        \State Compute valid length $\ell_i$, return $R_i=\sum_{k=0}^{H-1}m_{i,k}\gamma^k r_{i+k}$, and terminal flag $d_i$
        \State Assemble the next normalized base chunk $\bar{a}^{b}_{i+\ell_i:i+\ell_i+H}$ from replay
        \State Compute next corrected chunk $\bar{a}'_{i+\ell_i:i+\ell_i+H}$ from $\pi_{\theta'}$, $\bar{a}^{b}_{i+\ell_i:i+\ell_i+H}$, and target smoothing noise
        \State $\widehat{Q}_i\leftarrow\min_{j\in\mathcal{J}_2}Q_{\phi'_j}(s_{i+\ell_i},\bar{a}^{b}_{i+\ell_i:i+\ell_i+H},\bar{a}'_{i+\ell_i:i+\ell_i+H})$
        \State $y_i\leftarrow R_i+(1-d_i)\gamma^{\ell_i}\widehat{Q}_i$
        \State Update $\{\phi_j\}_{j=1}^{N_Q}$ and $\omega$ to minimize $\frac{1}{N_Q}\sum_{j=1}^{N_Q}\mathrm{MSE}(Q_{\phi_j}(s_i,\bar{a}^{b}_{i:i+H},\bar{a}_{i:i+H}),y_i)$
        \State Soft-update critic and encoder targets

        \If{past critic warmup and actor update step}
            \State $\bar{a}^{\theta}_{i:i+H}\leftarrow\mathrm{clip}(\bar{a}^{b}_{i:i+H}+\pi_\theta(s_i,\bar{a}^{b}_{i:i+H}),-1,1)$
            \State Update $\theta$ to maximize $\frac{1}{N_Q}\sum_{j=1}^{N_Q} Q_{\phi_j}(s_i,\bar{a}^{b}_{i:i+H},\bar{a}^{\theta}_{i:i+H})$ with residual L2 regularization
            \State Soft-update actor target
        \EndIf
    \EndFor
\Until{convergence}
\end{algorithmic}
\end{algorithm}

\subsection{Evaluation and Results}
\label{app:dpp-rl-result}

We evaluate in the same simulator and anchor-cell set used for training, with exploration noise disabled.
The primary metric is task success rate.
Evaluation uses an anchor-balanced protocol over the four predefined anchor cells.
For residual RL, we evaluate checkpoints from four training seeds at the same training step.
Each residual RL seed uses 20 episodes per anchor.
\cref{tab:aux-rl-anchor-balanced} summarizes the anchor-balanced evaluation results.

\begin{table}[h]
\centering
\caption{Anchor-balanced evaluation results for the auxiliary residual RL experiment. 
Success rates are averaged over four seeds; residual RL is evaluated at 230k checkpoint.}
\label{tab:aux-rl-anchor-balanced}
\begin{tabular}{lc}
\toprule
Policy & Success rate \\
\midrule
Base policy only
& $52.2 \pm 7.2\%$ \\
Residual RL
& $74.7 \pm 1.6\%$ \\
\bottomrule
\end{tabular}
\end{table}

At this checkpoint, residual RL improves the mean success rate from $52.2\%$ to $74.7\%$.
We view this as evidence that the base policy can serve as a useful prior for interaction-based residual adaptation.
The result is still auxiliary: online RL is not monotonic, and performance remains sensitive to seed and checkpoint choice.

\begin{figure}[t]
    \centering
    \includegraphics[width=\linewidth]{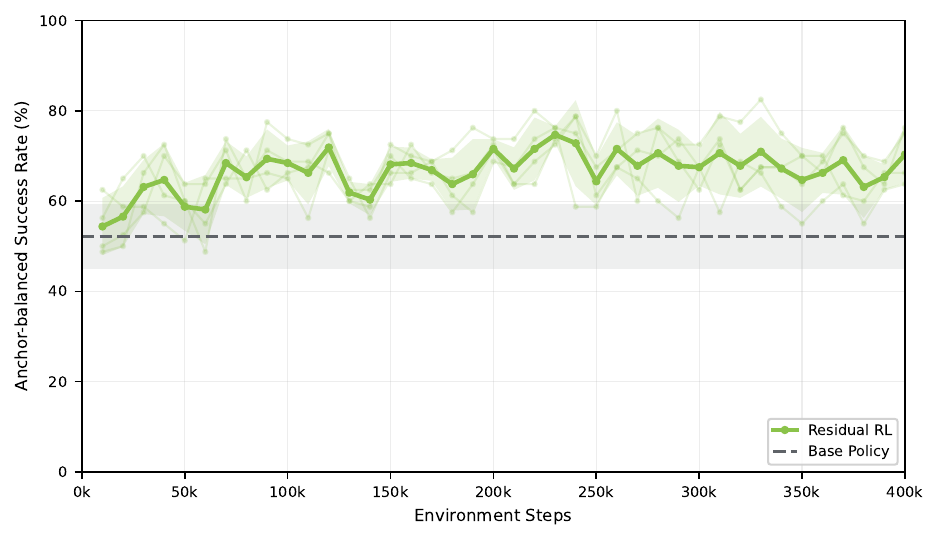}
    \caption{\textbf{Anchor-balanced success rate.}
    Residual RL improves the base-policy reference under the same anchor-balanced evaluation protocol.}
    \label{fig:dpp-rl-success-curve}
\end{figure}

\cref{fig:dpp-rl-success-curve} shows the checkpoint-wise evaluation curve for residual RL alongside the base-policy-only reference. The results indicate that residual RL improves over the frozen base policy, although its performance is non-monotonic across training.

\begin{figure}[t]
    \centering
    \begin{minipage}{0.48\linewidth}
        \centering
        \includegraphics[width=\linewidth]{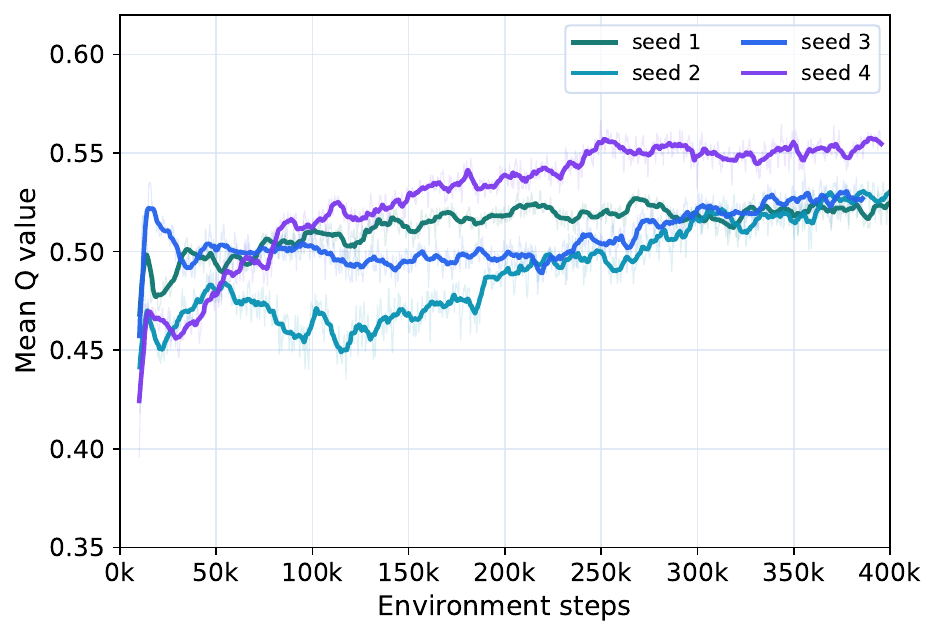}
        \vspace{0.2em}
        (a) Mean Q value
    \end{minipage}
    \hfill
    \begin{minipage}{0.48\linewidth}
        \centering
        \includegraphics[width=\linewidth]{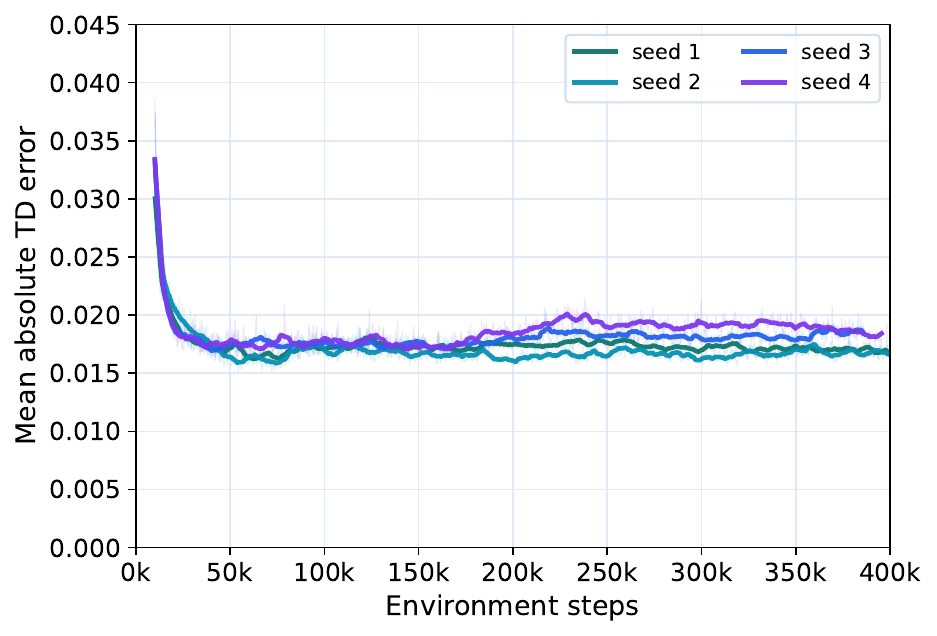}
        \vspace{0.2em}
        (b) TD error
    \end{minipage}
    \caption{\textbf{Residual RL training diagnostics.}
    Bounded Q values and TD errors indicate stable chunk-level residual RL optimization.}
    \label{fig:aux-rl-training-diagnostics}
\end{figure}

\paragraph{Training diagnostics.}
We report Q values and TD errors as auxiliary diagnostics for residual RL training.
\cref{fig:aux-rl-training-diagnostics} overlays these quantities across four seeds up to 400k environment steps.
The bounded curves suggest that the chunk-level Bellman updates remain numerically well behaved in this setting.
Task success remains the primary evaluation metric.

\subsection{Interpretation and Limitations}
\label{app:dpp-rl-discussion}

These results suggest that, in this simulation setting, the dexterous point policy can serve as a useful prior for residual RL.
The base policy already provides object-centric hand trajectories and a contact prior, allowing the residual policy to focus on bounded local corrections.
The Q-chunk critic is also compatible with this setting because the base policy produces action chunks.

We emphasize that this is an auxiliary simulation study.
It is limited to the spherical-object-to-bowl task and the four-cell anchor-balanced evaluation distribution, and it does not modify the main human-video-only training pipeline.
Overall, this auxiliary study provides preliminary evidence for the feasibility of residual RL adaptation on top of the base policy in simulation.

\subsection{Residual RL Formulation and Environment-Specific Settings}
\label{app:dpp-rl-settings}

\cref{tab:dpp-rl-formulation} summarizes the residual RL formulation and simulator configuration, \cref{tab:dpp-rl-anchor-cells} lists the anchor-cell reset distribution, and \cref{tab:dpp-rl-hparams} reports the hyperparameters used in the auxiliary residual RL experiment.

\begin{table}[t]
\centering
\caption{Residual RL formulation and environment-specific settings.}
\label{tab:dpp-rl-formulation}
\begin{tabular}{l|c}
\hline
Setting & Value \\
\hline
Action space & \texttt{kp18\_qchunk} \\
Single-step action dimension & 18 \\
Chunk horizon $H$ & 15 \\
Chunk action dimension & $15\times18=270$ \\
Replay storage & frame-level transitions \\
Chunk sampler & Q-chunking sequence sampler \\
Point encoder dimension & 128 \\
MLP hidden dimension & 1024 \\
Past contact input & enabled, 5D \\
Action range expansion & 0.2 \\
Max environment steps & 400,000 \\
\hline
\end{tabular}
\end{table}
\begin{table}[t]
\centering
\caption{Anchor-cell reset distribution for the spherical object. Coordinates are in meters.}
\label{tab:dpp-rl-anchor-cells}
\begin{tabular}{c|c|c}
\hline
Anchor & $x$ range & $y$ range \\
\hline
a1 & $[0.290, 0.310]$ & $[-0.120, -0.100]$ \\
a2 & $[0.370, 0.390]$ & $[-0.160, -0.140]$ \\
a3 & $[0.270, 0.290]$ & $[-0.120, -0.100]$ \\
a4 & $[0.330, 0.350]$ & $[-0.160, -0.140]$ \\
\hline
\end{tabular}
\end{table}
\begin{table}[t]
\centering
\caption{RL training hyperparameters used for the reported residual RL experiments.}
\label{tab:dpp-rl-hparams}
\begin{tabular}{l|c}
\hline
Parameter & Value \\
\hline
Discount $\gamma$ & 0.995 \\
Replay capacity & 200,000 \\
Offline / online batch ratio & 0.5 / 0.5 \\
Batch size & 256 \\
Q ensemble size $N_Q$ & 10 \\
Target Q subset size & 2 \\
Actor learning rate & $1\times10^{-6}$ \\
Critic learning rate & $1\times10^{-4}$ \\
Target update coefficient $\tau$ & 0.005 \\
Target noise / clip & 0.025 / 0.3 \\
Actor update frequency & 4 \\
Critic warmup updates & 10,000 \\
Residual policy scale $\alpha_{\mathrm{actor}}$ & 0.05 \\
Residual L2 penalty $\lambda_{\mathrm{res}}$ & $1\times10^{-4}$ \\
\hline
\end{tabular}
\end{table}


\end{document}